\expandafter\ifx\csname ifORsubmission\endcsname\relax
  \expandafter\newif\csname ifORsubmission\endcsname % false unless main_or.tex set it
\fi

\documentclass[11pt]{article}

\usepackage[margin=1in]{geometry}
\usepackage{parskip}
\usepackage{setspace}
\ifORsubmission
\else
\fi
\usepackage[T1]{fontenc}
\usepackage{microtype}

\usepackage{amsmath,amssymb,amsthm}
\ifORsubmission
  \usepackage{txfonts}% Times, as in the INFORMS template; load after the AMS packages
\else
  \usepackage[sc]{mathpazo}
\fi
\usepackage{bm}

\newtheorem{proposition}{Proposition}
\newtheorem{corollary}{Corollary}
\newtheorem{lemma}{Lemma}
\theoremstyle{definition}
\newtheorem{assumption}{Assumption}
\newtheorem{definition}{Definition}

\theoremstyle{remark}

\DeclareMathOperator{\Var}{Var}
\DeclareMathOperator{\Cov}{Cov}
\DeclareMathOperator{\Corr}{Corr}
\newcommand{\E}{\mathbb{E}}
\newcommand{\R}{\mathbb{R}}
\newcommand{\cx}{\succeq_{\mathrm{cx}}}

\usepackage{pgfplots}
\pgfplotsset{compat=1.18}
\usepackage{graphicx}
\usepackage{xcolor}
\usepackage{booktabs}
\usepackage{array}
\usepackage{enumitem}
\setlist{topsep=0.3em, itemsep=0.25em, parsep=0.1em, leftmargin=1.5em}
\usepackage{float}
\ifORsubmission
  \let\origtable\table \let\endorigtable\endtable
  \usepackage[nomarkers,nofigures,nolists]{endfloat}
  
\fi

\usepackage[round]{natbib}
\setcitestyle{aysep={}}
\usepackage[colorlinks=true, allcolors=blue!60!black]{hyperref}
\usepackage[capitalise,noabbrev]{cleveref}
\crefname{assumption}{Assumption}{Assumptions}
\crefname{example}{Example}{Examples}

\begin{document}

\begin{center}
{\LARGE\bfseries When Do Experiments Improve\\Treatment Prioritization?\par}
\ifORsubmission\else
\vspace{1.2em}
{\large Carlos Fern\'andez-Lor\'ia$^{1}$ \qquad Yanfang Hou$^{1}$ \qquad Jorge Lor\'ia$^{2}$\par}
\vspace{0.4em}
{\normalsize $^{1}$Hong Kong University of Science and Technology \qquad $^{2}$Aalto University\par}
\vspace{0.2em}
{\small\texttt{imcarlos@ust.hk}, \texttt{yhouav@connect.ust.hk}, \texttt{jorge.loria@aalto.fi}\par}
\vspace{0.6em}
{\today\par}
\fi
\end{center}
\vspace{1em}

\begin{abstract}
\noindent
Organizations often prioritize interventions using predictions of what would happen without them. We ask when experiments improve prioritization beyond these predictions. In a model of binary outcomes, predicted probabilities reveal whose outcome is uncertain, and so whom treatment can tip, which is why they often rival effect estimates. Yet they can fail in two ways: where effects peak along the predictions depends on the treatment's strength, and people with the same prediction may respond differently. An experiment can address both, at very different costs. Learning the treatment's strength takes only a small experiment and shows how to rank by the predictions; even ranking by closeness to 50\% often works well. Learning how people with the same prediction differ is harder, and the data needed depend on these differences, not on overall effect variation. When these differences are small, learning them improves allocation only if they outweigh estimation error near the cutoff that treatment capacity imposes on the ranking. In an advertising experiment with 14 million users, effect estimates improve narrow campaigns with a few thousand treated users but bring no detectable gain to broad ones even with 8 million.

\medskip
\noindent\textbf{Keywords:} treatment prioritization, heterogeneous treatment effects, value of experimentation

\ifORsubmission
\medskip
\noindent\textbf{Subject classifications:} Statistics: design of experiments: value of experimental data for treatment prioritization; Decision analysis: applications: targeting interventions with predictions; Marketing: segmentation: uplift modeling and causal targeting.

\medskip
\noindent\textbf{Area of review:} Machine Learning and Data Science.
\fi
\end{abstract}

\newpage

% Central idea: a good ranking needs potential, recovery, and monotonicity. Outcome predictions (the baseline) often have high potential and recovery but can fail monotonicity; an experiment can fix the order or learn what the baseline misses.

% Intro structure: two learning tasks (fixing the order; learning what the baseline misses). The five insights are collected in the Discussion (Table tab:map): (1) predictions are competitive for a reason; (2) information is not ordering; (3) fixing the order is cheap; (4) forgone, not total, heterogeneity; (5) value through decisions near the treatment cutoff.

% Evidence: running-example simulations and the Criteo experiment, where CATE prioritization wins in narrow campaigns with a few thousand treated users but brings no detectable gain in broad ones even with 8 million.

% =====================================================================
\section{Introduction}
% =====================================================================

Organizations with limited resources often need to decide whom to prioritize. A company might offer discounts to retain customers, a hospital might provide additional care to prevent readmissions, or an advertiser might choose whom to show ads. A common approach is to rank people by predictions of what's likely to happen without the intervention, which we call \textbf{baseline predictions}. But the people most likely to leave, return to hospital, or buy are not necessarily those whose outcomes we can change most.

One alternative is to rank people by their estimated \textbf{conditional average treatment effects (CATEs)}: the expected difference between outcomes with and without treatment, given their characteristics. Although this approach directly targets the treatment effect, the evidence on whether it improves targeting is mixed. \citet{ascarza2018} finds that the customers most likely to churn are not those most responsive to retention efforts and argues for targeting based on estimated effects instead. Other studies also find gains from doing so \citep{hitsch2024heterogeneous, yang2024targeting}. Yet these estimates can be too noisy to improve targeting \citep{radcliffe2011}. In a financial-aid experiment, a logit combining baseline predictions with one treatment coefficient outperformed a causal forest \citep{athey2025}. Causal-classification work explains how baseline predictions can outperform effect estimates through a bias--variance trade-off \citep{fernandez2022causal, fernandez2022decision}.

\emph{When can experiments improve treatment prioritization beyond what baseline predictions already provide?} An experiment can help ``fix the order'' by showing us which baseline predictions are associated with the largest average treatment effects. It can also help us learn what the baseline misses by revealing differences in treatment effects among people with the same baseline. We show why fixing the order can require much less experimental data. The value of learning what the baseline misses depends on whether we can estimate those differences accurately enough to improve decisions about whom to treat. Treatment capacity sets the cutoff in the ranking between those we treat and those we leave untreated. When the differences are small, they can change who gets treated only near this treatment cutoff. Following \citet{geoffrion1976purpose}, we use a tractable model to understand these forces, then examine them in simulations and a large advertising experiment.

For binary outcomes, such as churn or purchases, a \textbf{threshold model} connects baseline predictions to treatment effects: the outcome happens when an underlying index crosses a threshold, and treatment shifts that index. The index may represent utility, as in models of choice \citep{mcfadden1974conditional}, but it need not. It has two parts: a systematic index, determined by a person's characteristics, and a random component. Because of the random component, we cannot predict the outcome with certainty, but we can estimate its probability. The probability without treatment is the \textbf{baseline}, and the \textbf{effect} is how much treatment changes it.

\begin{figure}[t]
\centering
\begin{tikzpicture}
  \begin{axis}[
    width=9.5cm, height=5.8cm,
    axis lines=left,
    xlabel={Systematic index},
    ylabel={Probability},
    ytick={0,0.25,0.5,0.75,1},
    domain=-5:5, samples=200, smooth,
    ymin=0, ymax=1.02,
    grid=major, grid style={gray!15},
    legend pos=north west, legend cell align=left, legend style={font=\small, draw=none, fill=none},
    every axis plot/.append style={line width=1.3pt}
  ]
    \addplot[blue!70!black] {1/(1+exp(-x))};
    \addlegendentry{Baseline}
    \addplot[orange!80!black, dashed] {1/(1+exp(-x-0.5)) - 1/(1+exp(-x))};
    \addlegendentry{Effect}
    \draw[gray, dotted, line width=0.8pt] (axis cs:-0.25,0) -- (axis cs:-0.25,0.85);
    \node[gray!40!black, font=\small, anchor=south, inner sep=2pt] at (axis cs:-0.25,0.85) {effect peak};
    \addplot[only marks, mark=*, mark size=1.8pt, orange!80!black, forget plot] coordinates {(-0.25,0.1244)};
  \end{axis}
\end{tikzpicture}
\caption{\textbf{The baseline and the effect are functions of one index.} Here the constant treatment shift is 0.5, and the random component of the index follows a logistic distribution. The effect peaks half a shift below the outcome threshold (dotted line).}
\label{fig:hump}
\end{figure}

\Cref{fig:hump} shows the simplest case, in which treatment shifts everyone's index by the same amount. Three things stand out. First, how much treatment can change someone's outcome depends on how uncertain that outcome is, which the baseline captures: people with the same baseline have the same effect, so the effect is a function of the baseline. Second, effects vary much less than baselines in this example, so estimation error that barely disturbs a ranking by baselines can scramble a ranking by effects. Third, the baseline orders effects on either side of the effect peak, but not across it. Left of the peak, higher baselines mean larger effects, but right of it, the reverse relationship holds. 

Together, these observations explain why baseline predictions can compete with effect estimates and why there is still room to improve on them. Baselines are often easier to estimate, especially when untreated data are abundant, and can be highly informative about treatment effects. Yet even when the baseline captures every difference in effects, ranking people by it can still put the wrong people first. Beyond this example, people with the same baseline may also respond differently to treatment. An experiment can help with both problems.

\emph{Fixing the order.} Consider a nudge, such as a reminder. The people most likely to act may need no reminder, while those least likely may ignore it. We might then prioritize people whose baseline is closest to 50\%, an approach we call \textbf{threshold prioritization}, which requires no experiment. But effects tend to peak somewhat below 50\%, and further below the stronger the nudge, so threshold prioritization is just an approximation for cases with modest nudges. An experiment can do better by estimating the \textbf{baseline-conditioned average treatment effect (BATE)}, the average effect among people with the same baseline. Ranking by the BATE, which we call \textbf{BATE prioritization}, makes the best possible use of the baseline.

\emph{Learning what the baseline misses.} Even people with the same baseline can differ in how much treatment would help them. We call this variation \textbf{forgone heterogeneity}. Once we make the best use of the baseline, any further gain must come from learning these differences. Effects can vary widely and still be fully captured by the baseline, for instance when customers who care more about price are both less likely to renew and more responsive to a discount. Ranking by estimated effects, \textbf{CATE prioritization}, captures forgone heterogeneity but adds estimation error, and estimating it often takes far more data than the BATE. When these differences are small, learning them can change whom we treat only near the cutoff. Whether doing so improves prioritization depends on how large those differences are there relative to the error we introduce by estimating them. Since treatment capacity sets the cutoff, the same experiment can be large enough to improve prioritization at one capacity and too small at another.

\Cref{tab:ladder} summarizes the four rankings and when each works best.

\begin{table}[t]
\centering
\caption{\textbf{Four rankings.} Baseline and threshold prioritization need no experiment; threshold prioritization uses theory to fix the order. BATE prioritization uses an experiment to fix the order, and CATE prioritization uses one to learn what the baseline misses, at the cost of noisier estimates. The last column describes when each ranking works best under a threshold model.}
\label{tab:ladder}
\vspace{4pt}
\small
\renewcommand{\arraystretch}{1.25}
\begin{tabular}{@{}>{\raggedright\arraybackslash}p{2.1cm} >{\raggedright\arraybackslash}p{3.6cm} >{\raggedright\arraybackslash}p{2.6cm} >{\raggedright\arraybackslash}p{6.2cm}@{}}
\toprule
Ranking & Ranks people by & Needs & Works best when \\
\midrule
Baseline & Their baseline or its reverse & Untreated data & Treatment shifts everyone equally, all baselines on one side of the effect peak \\
Threshold & How close their baseline is to 50\% & Untreated data & Treatment shifts everyone equally and modestly \\
BATE & Average effect given baseline & An experiment & Effects depend on individual characteristics only through the baseline \\
CATE & Their estimated effect & An experiment, often a large one & Effect estimates are sufficiently accurate \\
\bottomrule
\end{tabular}
\end{table}

The distinction between these two tasks helps explain why studies comparing baseline predictions with estimated treatment effects can reach different conclusions. An experiment may improve targeting simply by showing how to order people using their baselines. How much more it can achieve depends on what those baselines miss and where in the ranking those differences matter. We use an advertising experiment with 14 million users to illustrate how much the answer can change even within a single setting. Among the top 1\%, CATE prioritization beats BATE prioritization with just a few thousand treated users and nearly triples its value with a million. Among the top half, it brings no detectable gain even with 8.3 million treated users.

% =====================================================================
\section{Treatment Prioritization}\label{sec:setup}
% =====================================================================

Let $X$ denote a person's observed characteristics, $T\in\{0,1\}$ the treatment, and $Y(1)$, $Y(0)$ binary potential outcomes with and without treatment. A person's \textbf{baseline} is their probability of the event of interest without treatment, and their \textbf{effect} is the CATE:
\begin{equation}
  \theta(x)=\mathbb P(Y(0)=1\mid X=x),
  \qquad
  \beta(x)=\mathbb P(Y(1)=1\mid X=x)-\theta(x).
\end{equation}
Where precision matters, we call $\theta$ the \textbf{baseline probability} and $\theta_1=\theta+\beta$ the \textbf{treated probability}.

A \textbf{score} $S=S(X)$ ranks people, with higher scores treated first. We consider treatments that increase the outcome probability. For interventions intended to prevent an outcome, such as readmission, the results apply to its complement.

The rankings in \cref{tab:ladder} use this information differently. Baseline prioritization ranks by $\theta$ or its reverse; threshold prioritization by $-|\theta-1/2|$; and BATE prioritization by $m_\theta=\E[\beta\mid\theta]$, the best use of the baseline (\cref{sec:lens}). CATE prioritization ranks by an estimate $\hat\beta$ of the effect. Ranking names refer to their true targets unless qualified as estimated, except for CATE prioritization, which always uses estimated effects: the true CATE gives the optimal ranking by definition.

\subsection{Measuring Prioritization}

We compare rankings at a given \textbf{capacity} $u\in(0,1]$, the fraction of the population that can be treated, as set by a budget or other constraints; we take it as given and do not study how to choose it. With equal treatment costs and equal value for each additional outcome, the objective is the cumulative effect. For a continuously distributed score $S$ with CDF $F_S$, treating its top fraction $u$ yields an expected value of
\begin{equation}\label{eq:curve}
  V(u;S)=\E\big[\beta(X)\,\mathbf 1\{F_S(S)\ge 1-u\}\big],
\end{equation}
per member of the population. In the presence of ties, as in discrete cases, people at the boundary are chosen at random.

We call $u\mapsto V(u;S)$ the score's \textbf{prioritization curve}. Ranking by $\beta$ itself maximizes $V(u;\cdot)$ at every $u$. Because treatment capacity may be unknown when a ranking is built, we compare scores across all treatment fractions.

\begin{definition}[Uniform dominance]\label{def:dominance}
Score $S_1$ \emph{uniformly dominates} $S_2$ if $V(u;S_1)\ge V(u;S_2)$ for every $u\in(0,1]$.
\end{definition}

Uniform dominance is demanding. It implies dominance under every rank-weighted average treatment effect with nonnegative weights, including AUTOC and the Qini coefficient \citep{yadlowsky2025rate}. When prioritization curves cross, which ranking is better depends on how many people can be treated, a theme we return to in \cref{sec:narrow}.

We also compare rankings by their \textbf{average performance} across treatment fractions,
\begin{equation}\label{eq:average-value}
A(S)=\int_0^1 V(u;S)\,du.
\end{equation}
Random targeting has $V(u)=u\,\E[\beta]$ and so $A=\E[\beta]/2$, which accounts for most of what even the best ranking achieves. To focus on the gains from prioritization, we normalize by subtracting $\E[\beta]/2$ and dividing by the gain achieved by the best ranking:
\begin{equation}\label{eq:qini-share}
\frac{A(S)-\E[\beta]/2}{A(\beta)-\E[\beta]/2}.
\end{equation}
We call this measure \textbf{normalized Qini}, since the numerator is the Qini coefficient of \citet{yadlowsky2025rate}. Random targeting scores zero and the best ranking one. We report average performance as normalized Qini throughout the paper.

\subsection{Potential, Recovery, and Monotonicity}\label{sec:lens}

A score can fail in three ways: it can miss information about effects, lose that information in estimation, or put people in the wrong order. To distinguish these failures, let $m_S=\E[\beta\mid S]$ be the expected effect among people with the same score. The share of effect variance the score explains is
\begin{equation}
  R^2_{\beta\mid S}=\frac{\Var(m_S)}{\Var(\beta)}.
\end{equation}
Let $q$ be a score's target and $\hat q$ its estimate; for the baseline, $q=\theta$, and for an effect estimate, $q=\beta$. 
In the next proposition we decompose the share of variance when the estimation procedure of the score is included.

\begin{proposition}[Potential and recovery]\label{prop:decomp} Suppose estimation adds no information about effects beyond the target: $\E[\beta\mid\hat q,q]=\E[\beta\mid q]$. Then
\begin{equation}
  R^2_{\beta\mid\hat q}=\underbrace{R^2_{\beta\mid q}}_{\text{potential}}\cdot\underbrace{\rho_q^2}_{\text{recovery}},\qquad \rho_q=\Corr(m_q,m_{\hat q})\in[0,1].
\end{equation}
\end{proposition}

The first factor is \textbf{information potential}, or simply \textbf{potential}: how much effect variation the target reveals. When $R^2_{\beta\mid q}=1$, we say there is \textbf{complete potential}, and when $R^2_{\beta\mid q}<1$ we say it is incomplete. The second is \textbf{information recovery}, or \textbf{recovery}: how much survives estimation. The baseline has incomplete potential but is usually recovered well (\cref{sec:easier}). The first three rankings all draw on the baseline. Baseline and BATE prioritization keep its potential, while threshold prioritization can give up some by giving baselines $p$ and $1-p$ the same score. Their recovery is usually high but not identical, since BATE prioritization must also learn how effects vary with the baseline. CATE prioritization targets effects themselves, so its potential is complete, and its explained variance is all recovery, $R^2_{\beta\mid\hat\beta}=\rho_\beta^2$, which depends on the experiment.

Information alone does not determine the order. Using only what $S$ reveals, the best ranking puts people with the largest expected effects $m_S$ first. We call this \textbf{signal monotonicity}, or \textbf{monotonicity}: $m_S$ is nondecreasing in $S$, so higher scores mean larger expected effects. A one-to-one transformation preserves information but can reverse an order. Thus an informative baseline is not enough; we must also know, assume, or learn how effects vary with it.

Baseline prioritization assumes that effects move in one direction with the baseline. Threshold prioritization uses the ordering suggested by effects peaking near 50\%. BATE prioritization ranks by $m_\theta$ itself and is signal monotone by construction.

\begin{proposition}[Explained variance and prioritization]\label{prop:bar}
Let $S_1$ satisfy signal monotonicity.
\begin{enumerate}[label=(\alph*)]
  \item If $S_2$ carries no information about effects beyond $S_1$, that is, $\E[\beta\mid S_1,S_2]=\E[\beta\mid S_1]$, then $S_1$ uniformly dominates $S_2$, and $R^2_{\beta\mid S_1}\ge R^2_{\beta\mid S_2}$.
  \item If $S_2$ also satisfies signal monotonicity and the CDFs of $m_{S_1}$ and $m_{S_2}$ cross at most once, then $S_1$ uniformly dominates $S_2$ if and only if $R^2_{\beta\mid S_1}\ge R^2_{\beta\mid S_2}$.
\end{enumerate}
\end{proposition}

Part (a) makes BATE prioritization the gold standard of baseline-based rankings. Baseline and threshold prioritization approximate this ideal under different conditions.

Part (b) covers scores that draw on different information, such as an estimated baseline and an estimated effect. Here explained variance can mislead. One score may single out the few people with the very largest effects, while another sorts everyone else better; which is preferable then depends on how many we can treat. Single crossing rules this out. The condition says that one score's expected effects are lower among the people it ranks lowest and higher among those it ranks highest: it spreads people apart at both ends. A score that spreads people apart more is better at every capacity, and explained variance measures that spread. When the condition fails, we compare allocations capacity by capacity (\cref{sec:narrow}).

When both rankings satisfy signal monotonicity and the distributions of their expected effects cross at most once, part (b) and \cref{prop:decomp} give a simple test for CATE prioritization. An effect estimate $\hat\beta$ uniformly dominates the best ranking based on an estimated baseline $\hat\theta$ if and only if
\begin{equation}\label{eq:bar}
  \rho_\beta^2\;\ge\;R^2_{\beta\mid\theta}\,\rho_\theta^2.
\end{equation}
The left side is the effect estimate's recovery, since its potential is complete; the right side is the baseline's potential times its recovery. A ranking with complete potential can therefore still lose when the baseline explains much of the variation in effects and is estimated more reliably.

\subsection{Related Work}\label{sec:related}

Baseline and CATE prioritization are both established approaches. Customer management has a long tradition of ranking by predicted risk \citep{neslin2006defection}, alongside work targeting estimated effects \citep{ascarza2018, simester2020targeting, yoganarasimhan2023design, hitsch2024heterogeneous, yang2024targeting}, using estimators such as those of \citet{kunzel2019metalearners} and \citet{nie2021quasi}. Clinical research distinguishes risk modeling---examining benefit across strata of baseline risk, a prognostic score \citep{hansen2008prognostic}---from effect modeling, which estimates treatment-by-covariate interactions directly. The latter can overstate heterogeneity and mistarget treatment \citep{kent2007limitations, vanklaveren2019models, kent2020path}. A common shift in our logistic threshold model corresponds to a constant odds ratio, the standard assumption in risk modeling.

\citet{athey2025} provide the closest connection to our work. In a model of filing decisions, targeting intermediate baselines works best, and adding a single treatment coefficient to predicted baseline log-odds outperforms a causal forest. They attribute this to noise in nonparametric effect estimates. We explain why learning this correction can require little data, while recovering effect differences among people with the same baseline can require much more. Separating these two learning tasks identifies when recovering additional information improves allocation: locally, forgone heterogeneity must outweigh estimation error at the treatment cutoff, so gains at one capacity need not extend to another.

The causal-classification analyses of \citet{fernandez2022causal, fernandez2022decision} show that good treatment decisions need not require accurate effect estimates and that bias in outcome predictions can help those decisions. We study how these forces determine the experimental data needed to improve on an existing baseline ranking at each treatment capacity. Rank-weighted average treatment effects evaluate prioritization rules across treatment fractions \citep{yadlowsky2025rate}, and \citet{chernozhukov2025generic} compare machine-learning proxies for treatment effects by how much effect variation their best linear predictor explains, a linear analogue of the explained variance we use. When the goal is to select the units whose true effects are in the top fraction, the best selection rule depends on that fraction \citep{gu2023invidious}. We study instead the total effect of those treated, where capacity matters through the people near the cutoff.

Our comparison of scores builds on a classic idea: a more informative signal produces more spread-out expected values \citep{blackwell1953equivalent, ganuza2010signa}, which the convex order formalizes \citep{shaked2007stochastic}. \Cref{prop:bar}(a) applies this idea to prioritization. Part (b) establishes when explained variance is enough to compare rankings, connecting potential and recovery to prioritization performance.

More broadly, prescriptive analytics seeks predictions that lead to good actions \citep{bertsimas2020predictive, elmachtoub2022smart}. Policy learning optimizes treatment rules directly \citep{kitagawa2018should, athey2021policy}, and \citet{manski2016sufficient} size trials for near-optimal treatment choices rather than for detecting effects. We ask what determines the experimental data needed to improve prioritization beyond existing predictions, and how the answer depends on treatment capacity.

\Cref{sec:threshold} introduces the threshold model and running example. \Cref{sec:enough} studies the first learning task: fixing the order. \Cref{sec:size} studies the second: learning what the baseline misses, at a given capacity and with noisy estimates. \Cref{sec:criteo} illustrates the mechanisms in advertising. Proofs are in \cref{app:proofs}.

% =====================================================================
\section{The Threshold Model}\label{sec:threshold}

The threshold model is our lens for the two learning tasks. It produces effects that rise and fall with the baseline, while allowing treatment shifts that the baseline may or may not reveal. We use that structure to explain the mechanisms at play, then examine their implications in simulations and an advertising experiment (\cref{sec:criteo}). We formalize the model here.
% =====================================================================

\begin{assumption}[Threshold model]\label{ass:threshold}
There are real-valued functions $\mu(x)$ and $\delta(x)$ such that
\begin{equation}
  Y(0)=\mathbf 1\{\mu(X)+\varepsilon>0\},\qquad Y(1)=\mathbf 1\{\mu(X)+\delta(X)+\varepsilon>0\},
\end{equation}
where $\varepsilon$ is independent of $X$ and has CDF $F$ and a continuous density $f(z)$ that is symmetric about zero and strictly decreasing in $|z|$.
\end{assumption}

We call $\mu(X)$ the \textbf{systematic index}, the part of the index determined by a person's characteristics ($X$), and $\delta(X)$ the \textbf{shift}. Logistic and Gaussian noise both satisfy the assumption; they correspond to the logit and probit links, and we refer to $F$ as the link. By symmetry,
\begin{equation}
  \theta=F(\mu),\qquad \beta=F(\mu+\delta)-F(\mu).
\end{equation}

\subsection{Why the Baseline Has High Potential}\label{sec:informative}

Imagine friends lounging by a swimming pool and a mischievous guest with one chance to push someone in. Who should it be? Someone standing at the edge needs only a light tap, while someone sitting several feet away would need a real shove. But someone near the edge may be about to jump in anyway, and pushing that person accomplishes nothing. The aim is to get someone wet who would otherwise have stayed dry.

In the model, the systematic index is closeness to the edge, the shift is the strength of the push, and the random component captures what the would-be shover cannot predict. The baseline is the chance of ending up in the water without a push; the effect is the chance that the push makes the difference. If everyone receives the same push, closeness to the edge is all that distinguishes their effects. This is the \textbf{common shift}, $\delta(x)=\delta$, in \cref{fig:hump}. In the next proposition, we characterize some conditions under which the baseline has complete potential and give an explicit form for $\beta$.

\begin{proposition}[Complete potential]\label{prop:sufficiency}
Under \cref{ass:threshold} with a common shift $\delta>0$, $\beta=h(\theta)$ with $h(p)=F(F^{-1}(p)+\delta)-p$. The function $h$ increases on $(0,\theta^*]$ and decreases on $[\theta^*,1)$, where $\theta^*=F(-\delta/2)$, and $R^2_{\beta\mid\theta}=1$.
\end{proposition}

The baseline therefore has complete potential, yet can still incorrectly order people. Moving closer to the edge makes a push more likely to work, but also makes an unprompted jump more likely. These forces balance half a push from the edge, at $\mu=-\delta/2$, where the effect peaks at $\theta^*$, below 50\%.

Left of the effect peak in \cref{fig:hump}, the \textbf{rising side}, higher baselines mean larger effects; rare outcomes such as conversions fall here. Right of the peak, the \textbf{falling side}, higher baselines mean smaller effects; common outcomes such as renewals fall here. If all baselines lie on the same side of the peak, baseline prioritization in the matching direction satisfies signal monotonicity, so no other ranking can do better. \Cref{sec:enough} examines alternatives when this condition fails.

Different pushes need not undermine this. If their strength depends only on proximity to the edge, the baseline still reveals everything. It loses potential only when strength varies among people equally close to the edge. Recovering this variation is the second learning task (\cref{sec:size}).

\subsection{Why the Baseline Is Recovered Well}\label{sec:easier}

\Cref{fig:hump} also shows that effects vary much less than baselines. An effect is the difference between two nearby probabilities, while the baseline is a probability itself. Seen as an estimate of effects, the baseline is therefore biased. But that bias may actually \emph{help} with the ranking.

Write a score's target as $q=\beta+\alpha$, the effect plus bias. The bias is \textbf{aligned} if it never decreases as effects increase. It then stretches the gaps between higher- and lower-effect people, so that estimation error is less likely to reverse their order (\cref{fig:stretch}). In the next proposition we study how the alignment of the bias can help decision making.

% Logistic noise, common shift 0.5. Person j: baseline 10%, effect 5.48 pts.
% Person i: baseline 20%, effect 9.19 pts. Same Gaussian noise, sd 3 pts.
% P(order reversed) = Phi(-gap/(3*sqrt 2)): 0.191 for effects, 0.0092 for baselines.
\begin{figure}[t]
\centering
\pgfmathdeclarefunction{npdf}{2}{\pgfmathparse{exp(-((x-#1)^2)/(2*9))/(3*sqrt(2*pi))}}
\begin{minipage}[t]{0.49\textwidth}
\centering
\begin{tikzpicture}
\begin{axis}[
    width=\linewidth, height=5.2cm,
    title={\small (A) Ranking by estimated effects},
    axis x line=bottom, hide y axis, xmin=-4, xmax=31, ymin=0, ymax=0.19,
    xlabel={Score (percentage points)},
    domain=-4:31, samples=120,
    tick label style={font=\scriptsize}, label style={font=\small},
    every axis plot/.append style={line width=1.1pt}
]
\addplot[orange!80!black] {npdf(5.4828,0)};
\addplot[blue!70!black, densely dashed] {npdf(9.1875,0)};
\draw[gray!70, dotted, line width=0.7pt] (axis cs:5.4828,0) -- (axis cs:5.4828,0.150);
\draw[gray!70, dotted, line width=0.7pt] (axis cs:9.1875,0) -- (axis cs:9.1875,0.150);
\draw[<->, >=stealth, black!70, line width=0.5pt] (axis cs:5.4828,0.150) -- (axis cs:9.1875,0.150);
\node[font=\scriptsize, anchor=base] at (axis cs:7.335,0.157) {$\Delta\beta=3.7$};
\node[font=\scriptsize, orange!65!black, anchor=east] at (axis cs:1.6,0.112) {$j$};
\node[font=\scriptsize, blue!65!black, anchor=west] at (axis cs:12.9,0.112) {$i$};
\node[font=\scriptsize, align=left, anchor=north west] at (axis cs:16.5,0.175) {order reversed\\19\% of the time};
\end{axis}
\end{tikzpicture}
\end{minipage}\hfill
\begin{minipage}[t]{0.49\textwidth}
\centering
\begin{tikzpicture}
\begin{axis}[
    width=\linewidth, height=5.2cm,
    title={\small (B) Ranking by estimated baselines},
    axis x line=bottom, hide y axis, xmin=-4, xmax=31, ymin=0, ymax=0.19,
    xlabel={Score (percentage points)},
    domain=-4:31, samples=120,
    tick label style={font=\scriptsize}, label style={font=\small},
    every axis plot/.append style={line width=1.1pt}
]
\addplot[orange!80!black] {npdf(10,0)};
\addplot[blue!70!black, densely dashed] {npdf(20,0)};
\draw[gray!70, dotted, line width=0.7pt] (axis cs:10,0) -- (axis cs:10,0.150);
\draw[gray!70, dotted, line width=0.7pt] (axis cs:20,0) -- (axis cs:20,0.150);
\draw[<->, >=stealth, black!70, line width=0.5pt] (axis cs:10,0.150) -- (axis cs:13.705,0.150);
\draw[<->, >=stealth, black!70, line width=0.5pt] (axis cs:13.705,0.150) -- (axis cs:20,0.150);
\draw[black!70, line width=0.5pt] (axis cs:13.705,0.144) -- (axis cs:13.705,0.156);
\node[font=\scriptsize, anchor=base] at (axis cs:11.85,0.157) {$\Delta\beta$};
\node[font=\scriptsize, anchor=base] at (axis cs:16.85,0.157) {$\Delta\alpha=6.3$};
\node[font=\scriptsize, orange!65!black, anchor=east] at (axis cs:6.1,0.112) {$j$};
\node[font=\scriptsize, blue!65!black, anchor=west] at (axis cs:23.9,0.112) {$i$};
\node[font=\scriptsize, align=left, anchor=north west] at (axis cs:-3.5,0.185) {order reversed\\1\% of the time};
\end{axis}
\end{tikzpicture}
\end{minipage}
\caption{\textbf{The baseline amplifies differences in effects.} Two people, $i$ and $j$, with baselines of 20\% and 10\% (logistic noise, $\delta=0.5$). Both scores are estimated with the same noise, a standard deviation of 3 points, and $\Delta$ denotes the difference between $i$ and $j$. (A) Effects differ by $\Delta\beta=3.7$ points, and noise reverses the order 19\% of the time. (B) Baselines differ by $\Delta\beta+\Delta\alpha=10$ points, where $\Delta\alpha$ is the difference in bias. Noise reverses the order 1\% of the time.}
\label{fig:stretch}
\end{figure}

\begin{samepage}
\begin{proposition}[Amplification]\label{prop:amplify}
Let $\hat q=q+e_q$ and $\hat\beta=\beta+e_\beta$, where the estimation errors $e_q$ and $e_\beta$ have the same conditional distribution given $(\beta,q)$. Then,
\begin{enumerate}[label=(\alph*)]
  \item if the bias $\alpha$ is aligned, $\hat q$ uniformly dominates $\hat\beta$.
  \item Under \cref{ass:threshold} with a common shift $\delta>0$, let $q=\theta$ on the rising side and $q=-\theta$ on the falling side, so that $q$ ranks by the baseline in the matching direction. The bias is aligned on the falling side for any noise distribution, and on the rising side if $f(z+\delta)\le2f(z)$ throughout the range of the systematic index.
\end{enumerate}
\end{proposition}
\end{samepage}

Part (a) applies beyond the threshold model. Bias alignment helps any score, not just the baseline. Earlier work showed that such bias can improve treatment assignment \citep{fernandez2022causal}, and we extend it here to treatment prioritization. For logistic noise, treatment at most doubling the odds of the outcome is sufficient for bias alignment over the entire rising side. Larger odds ratios can also satisfy the condition on a restricted range of baselines (\cref{app:amplify}).

The baseline also has an advantage in estimation noise. It can often be learned from abundant untreated data organizations already hold, while effects must come from much smaller experiments and are notoriously hard to estimate \citep{lewis2015unfavorable}. For small effects, a difference in means has about twice the variance of the control mean from the same balanced experiment, and four times that of an untreated sample of the same size. 

\subsection{A Running Example}\label{sec:example}

We use one running example to see how potential, recovery, and monotonicity shape prioritization.

As in \cref{fig:hump,fig:stretch}, the noise is logistic and the common shift is 0.5, so effects peak at a baseline of $F(-0.25)\approx44\%$. We consider three populations, each spread uniformly over an interval of the systematic index. Baselines range from 3\% to 20\% in the \textbf{rare} population, from 60\% to 95\% in the \textbf{common} population, and from 10\% to 80\% in the \textbf{straddling} population. The first two lie on one side of the peak, with no baseline between the peak and 50\%, so baseline, threshold, and BATE prioritization coincide. Only the straddling population poses an ordering problem: threshold prioritization gets close, but the peak is not exactly at 50\%.

To vary potential, we divide each population into 50 equally likely cells: 10 baseline groups of 5 cells each. Cells in a group share a baseline. Under a common shift they also share an effect, so the baseline's potential is complete. Varying their shifts creates differences the baseline cannot see, adding forgone heterogeneity in a controlled amount. When we do so, the five cells' shifts are spread evenly around 0.5, with standard deviation $0.5r$, so $r$ measures unpredictable shifts relative to the expected one.

Recovery depends on how much data we collect and how we use them. We measure an experiment's size $n$ by the number of people it treats. It assigns $n/50$ of them to each cell and draws binomial outcomes. Unless stated otherwise, we treat the baseline as known, as when abundant untreated data estimate it well. This isolates what we must learn from the experiment. \cref{app:learned} considers the case where the baseline must also be estimated. Assuming the baseline is known favors rankings based on experimental data because baseline estimation errors disturb baseline rankings less than effect rankings (\cref{sec:easier}), so experimental rankings need more data to be competitive when the baseline is estimated. The mechanisms and qualitative conclusions remain unchanged.

We use the experimental data in two ways. The \textbf{one-parameter logit}, following \citet{athey2025}, holds the calibrated baseline log-odds fixed and estimates a common shift. Under a common shift the model is correct and the BATE equals the CATE; when shifts vary, it is a restrictive approximation. We use it for estimated BATE prioritization. The \textbf{cell estimator} estimates each cell's effect separately, by its treated mean minus the baseline, or by a difference in means when the baseline must also be learned. We use it for CATE prioritization. It treats every estimated difference as signal, so it is a simple benchmark rather than the best estimator available; \cref{app:pooled} repeats the main analyses with an estimator that shrinks cell estimates toward the logit.

Reported prioritization curves average over 3{,}000 simulated experiments of each size. A reported crossover or required sample size is the smallest tested size that meets the stated criterion at every larger tested size. \Cref{app:simdetails} gives the estimation and evaluation details.

% =====================================================================

\section{Fixing the Order}\label{sec:enough}
% =====================================================================

The first learning task is to put existing information in the right order. Under a common shift, the baseline reveals all effects. On one side of the effect peak, ranking by the baseline in the matching direction is optimal. Across both sides, we should instead rank by distance from the peak. Throughout this section, we maintain \cref{ass:threshold} with a common shift $\delta>0$ and a known CDF $F$, isolating the cost of a wrong order.

\begin{samepage}
\begin{proposition}[Distance from the peak]\label{prop:onenumber}
Ranking by $-|F^{-1}(\theta)+\delta/2|$, the distance of the systematic index from the peak, achieves the best possible prioritization at every treatment fraction.
\end{proposition}
\end{samepage}

The distance here is measured on the index scale, $\mu=F^{-1}(\theta)$, rather than directly in baseline probabilities. But how accurately must we locate the peak? Consider the continuous version of the straddling population in \cref{fig:peak}A, where people are spread evenly along the index, and suppose we can treat half of them. The best ranking selects an interval centered at the true peak, $\mu=-0.25$. The two dots mark its endpoints, where people have the same effect.

% Logistic common shift 0.5; straddling population: uniform index on [F^{-1}(0.1),F^{-1}(0.8)] = [-2.1972,1.3863].
% Exact value integral: [log(1+exp(x+0.5))-log(1+exp(x))]/3.5835. Generated by scratchpad/peak/gen_peak.py.
% Both panels mark threshold prioritization: center error epsilon=0.25.
\begin{figure}[t]
\centering
\begin{minipage}[t]{0.49\textwidth}
\centering
\begin{tikzpicture}
\begin{axis}[
    width=\linewidth, height=6.3cm,
    title={\small (A) The loss is in the two small wedges},
    axis lines=left, xmin=-2.1972, xmax=1.3863, ymin=0, ymax=17.8,
    xlabel={Systematic index $\mu$}, ylabel={Effect (percentage points)},
    domain=-2.1972:1.3863, samples=150,
    tick label style={font=\scriptsize}, label style={font=\small},
    every axis plot/.append style={line width=1.1pt}
]
% People selected under both rankings.
\addplot[draw=none, fill=gray!16, domain=-0.8959:0.6459, samples=60]
 {100*(1/(1+exp(-x-0.5))-1/(1+exp(-x)))} \closedcycle;
% Equal-height rectangles represent the parts that cancel.
\addplot[draw=none, fill=orange!18] coordinates
 {(-1.1459,0) (-1.1459,10.2676159) (-0.8959,10.2676159) (-0.8959,0)} \closedcycle;
\addplot[draw=none, fill=blue!12] coordinates
 {(0.6459,0) (0.6459,10.2676159) (0.8959,10.2676159) (0.8959,0)} \closedcycle;
% Surplus forgone on the left and shortfall admitted on the right.
\addplot[draw=orange!80!black, line width=0.4pt, fill=orange!75] coordinates {(-1.1458797,10.2676159) (-1.1458797,10.2676159) (-1.1375464,10.3030260) (-1.1292131,10.3382638) (-1.1208797,10.3733258) (-1.1125464,10.4082083) (-1.1042131,10.4429078) (-1.0958797,10.4774206) (-1.0875464,10.5117433) (-1.0792131,10.5458722) (-1.0708797,10.5798036) (-1.0625464,10.6135342) (-1.0542131,10.6470601) (-1.0458797,10.6803780) (-1.0375464,10.7134841) (-1.0292131,10.7463750) (-1.0208797,10.7790471) (-1.0125464,10.8114968) (-1.0042131,10.8437205) (-0.9958797,10.8757148) (-0.9875464,10.9074761) (-0.9792131,10.9390009) (-0.9708797,10.9702856) (-0.9625464,11.0013268) (-0.9542131,11.0321210) (-0.9458797,11.0626646) (-0.9375464,11.0929543) (-0.9292131,11.1229866) (-0.9208797,11.1527580) (-0.9125464,11.1822652) (-0.9042131,11.2115047) (-0.8958797,11.2404731) (-0.8958797,10.2676159) (-1.1458797,10.2676159)};
\addplot[draw=blue!70!black, line width=0.4pt, fill=blue!55] coordinates {(0.6458797,10.2676159) (0.6458797,10.2676159) (0.6542131,10.2320371) (0.6625464,10.1962931) (0.6708797,10.1603877) (0.6792131,10.1243242) (0.6875464,10.0881063) (0.6958797,10.0517376) (0.7042131,10.0152216) (0.7125464,9.9785618) (0.7208797,9.9417618) (0.7292131,9.9048251) (0.7375464,9.8677552) (0.7458797,9.8305557) (0.7542131,9.7932300) (0.7625464,9.7557816) (0.7708797,9.7182140) (0.7792131,9.6805306) (0.7875464,9.6427350) (0.7958797,9.6048305) (0.8042131,9.5668206) (0.8125464,9.5287088) (0.8208797,9.4904983) (0.8292131,9.4521926) (0.8375464,9.4137950) (0.8458797,9.3753090) (0.8542131,9.3367378) (0.8625464,9.2980847) (0.8708797,9.2593531) (0.8792131,9.2205463) (0.8875464,9.1816675) (0.8958797,9.1427199) (0.8958797,10.2676159) (0.6458797,10.2676159)};
\addplot[black!75] {100*(1/(1+exp(-x-0.5))-1/(1+exp(-x)))};
\draw[gray!70, densely dotted, line width=0.7pt] (axis cs:-1.1459,10.2676159) -- (axis cs:0.8959,10.2676159);
\addplot[only marks, mark=*, mark size=1.6pt, black!75] coordinates {(-1.1459,10.2676159) (0.6459,10.2676159)};
\draw[gray!70, dotted, line width=0.7pt] (axis cs:-0.25,0) -- (axis cs:-0.25,13.7);
\draw[black!60, dashed, line width=0.7pt] (axis cs:0,0) -- (axis cs:0,13.7);
\draw[<->, >=stealth, black!70, line width=0.5pt] (axis cs:-0.25,13.25) -- (axis cs:0,13.25);
\node[font=\scriptsize, black!80, anchor=west] at (axis cs:0.03,13.25) {$d=0.25$};
\node[font=\scriptsize, gray!40!black, anchor=east] (tp) at (axis cs:-0.62,16.6) {true peak};
\node[font=\scriptsize, black!70, anchor=west] (gp) at (axis cs:0.22,16.6) {guessed peak};
\draw[->, >=stealth, gray!50!black, line width=0.5pt] (tp.east) -- (axis cs:-0.28,13.85);
\draw[->, >=stealth, black!70, line width=0.5pt] (gp.west) -- (axis cs:0.03,13.85);
\node[font=\scriptsize, orange!65!black, anchor=north] at (axis cs:-1.021,3.0) {dropped};
\node[font=\scriptsize, blue!65!black, anchor=north] at (axis cs:0.771,3.0) {added};
\end{axis}
\end{tikzpicture}
\end{minipage}\hfill
\begin{minipage}[t]{0.49\textwidth}
\centering
\begin{tikzpicture}
\begin{axis}[
    width=\linewidth, height=6.3cm,
    title={\small (B) Small center errors lose little value},
    axis lines=left, xmin=0, xmax=0.5, ymin=0, ymax=60,
    xlabel={Error in the peak, $d$}, ylabel={Loss in normalized Qini (points)},
    legend style={font=\scriptsize, at={(0.03,0.97)}, anchor=north west, draw=none, fill=none},
    legend cell align=left, grid=major, grid style={gray!15},
    tick label style={font=\scriptsize}, label style={font=\small},
    every axis plot/.append style={line width=1.1pt}
]
\addplot[black, dashed, line width=0.8pt, domain=0:0.5, samples=60] {228.703873*x^2};
\addlegendentry{Bound}
\addplot[blue!70!black] coordinates {(0.00,0.0000) (0.01,0.0149) (0.02,0.0594) (0.03,0.1329) (0.04,0.2350) (0.05,0.3651) (0.06,0.5227) (0.07,0.7072) (0.08,0.9184) (0.09,1.1553) (0.10,1.4178) (0.11,1.7050) (0.12,2.0167) (0.13,2.3521) (0.14,2.7108) (0.15,3.0920) (0.16,3.4954) (0.17,3.9205) (0.18,4.3663) (0.19,4.8329) (0.20,5.3190) (0.21,5.8247) (0.22,6.3488) (0.23,6.8913) (0.24,7.4512) (0.25,8.0280) (0.26,8.6214) (0.27,9.2305) (0.28,9.8550) (0.29,10.4939) (0.30,11.1471) (0.31,11.8136) (0.32,12.4932) (0.33,13.1849) (0.34,13.8886) (0.35,14.6031) (0.36,15.3283) (0.37,16.0637) (0.38,16.8083) (0.39,17.5621) (0.40,18.3238) (0.41,19.0936) (0.42,19.8702) (0.43,20.6538) (0.44,21.4431) (0.45,22.2378) (0.46,23.0379) (0.47,23.8420) (0.48,24.6504) (0.49,25.4617) (0.50,26.2762)};
\addlegendentry{Loss}
\addplot[only marks, mark=*, mark size=1.6pt, blue!70!black, forget plot] coordinates {(0.25,8.0280)};
\end{axis}
\end{tikzpicture}
\end{minipage}
\caption{\textbf{Small center errors exchange people with similar effects.} The continuous version of the straddling population (\cref{sec:example}). (A) Moving the center from $-0.25$ to zero while treating half the population; the two wedges are the loss. (B) Loss in normalized Qini against the bound of \cref{prop:secondorder}; the dot marks the error in panel A.}
\label{fig:peak}
\end{figure}

Suppose we place the peak at zero instead, an error of $d=0.25$. The selected interval shifts right, dropping the orange strip and adding the blue strip. The two wedges represent the cost of the error: the extra benefit lost on the left and the shortfall admitted on the right. Their combined area is proportional to the loss. For small errors, halving the error approximately halves both the widths and the heights of these wedges, reducing the loss to a quarter. The next proposition extends this argument to nonuniform populations and intervals reaching the edge of the support.

\begin{samepage}
\begin{proposition}[Errors in the peak cost little]\label{prop:secondorder}
Suppose the systematic index $\mu$ is supported on an interval, with a density continuous and positive in its interior and bounded above by $\bar g$. Let $L_\delta=\sup_z|f(z+\delta)-f(z)|$ be the largest absolute slope of the effect curve, and write $\mu^*=-\delta/2$ for the peak. For every center error $d\in\R$ and treatment fraction $u\in(0,1]$,
\begin{equation}\label{eq:peak-loss}
  0\le V(u;\beta)-V(u;-|\mu-(\mu^*+d)|)
  \le \bar gL_\delta\,d^2.
\end{equation}
\end{proposition}
\end{samepage}

The bound formalizes the geometry we just discussed: density $\bar g$ limits the number of people exchanged, and slope $L_\delta$ limits the difference in their effects. Small shifts flatten the effect curve, making a misplaced peak less costly.

\Cref{fig:peak}B plots the loss in normalized Qini as the error grows, against the bound, which holds at every treatment fraction and so also on average. The error of $0.25$ (marked by the dot) lowers normalized Qini by 8 percentage points.

\subsection{Locating the Effect Peak}\label{sec:locate}

We can locate the peak in two ways: guess it from the outcome threshold, or estimate it from an experiment. Threshold prioritization does the first. By symmetry, ranking by $-|\theta-1/2|$ is the same as ranking by $-|\mu|$, so it centers the ranking at $\mu=0$ instead of $\mu=-\delta/2$, an error of $d=\delta/2$. In \cref{fig:peak}, $\delta=0.5$, so this is exactly the error shown in panel A.

\begin{corollary}[Threshold prioritization]\label{cor:threshold}
Under the conditions of \cref{prop:secondorder}, threshold prioritization loses at most $\bar gL_\delta\,\delta^2/4$ in $V(u;\cdot)$ relative to the best ranking, at every treatment fraction.
\end{corollary}

Threshold prioritization therefore works especially well when shifts are modest. It is not guaranteed to beat baseline prioritization, but it has a useful advantage: it does not need to know where people sit relative to the peak. If everyone lies below the peak or above 50\%, it gives the same ranking as baseline prioritization in the optimal direction. If the population straddles the peak, it ranks people by their distance from a point that, for modest shifts, is close to the true peak.

Practitioners have arrived at a similar rule. The ``movable middles'' framework of the Marketing + Media Alliance recommends targeting consumers whose probability of buying a brand is in a middle range, roughly 20\% to 80\%, on the grounds that they respond most to its advertising \citep{mma2023movable}. Its rationale is a logit model of purchase, the threshold model with logistic noise. Our results clarify when this rule works: the effect peaks at a baseline probability close to 50\% when the treatment shift is modest. They also clarify what the rule misses: differences in treatment effects among people with the same baseline.

Threshold prioritization needs scores on a probability scale, but not full calibration. Scaling the predicted log-odds leaves its ranking unchanged, and shifting them moves the implied peak, an error that costs little.

\begin{figure}[t]
\centering
\begin{minipage}[t]{0.49\textwidth}
\centering
\begin{tikzpicture}
\begin{axis}[
    width=\linewidth, height=6cm,
    title={\small (A) Common shift},
    xmode=log, xmin=150, xmax=1.6e7, ymin=0, ymax=102,
    xlabel={Experiment size $n$ (treated people)}, ylabel={Normalized Qini (\%)},
    xtick={1e3,1e4,1e5,1e6,1e7}, xticklabels={$10^3$,$10^4$,$10^5$,$10^6$,$10^7$},
    legend columns=2, legend to name=straddleleg, legend style={font=\scriptsize, draw=none, /tikz/every even column/.append style={column sep=10pt}},
    legend cell align=left, grid=major, grid style={gray!15},
    every axis plot/.append style={line width=1.3pt}, tick label style={font=\scriptsize}, label style={font=\small}
]
\addplot[orange!80!black, densely dashed] coordinates {(150,98.80) (200,99.12) (250,99.26) (300,99.39) (350,99.44) (400,99.50) (450,99.53) (500,99.57) (550,99.61) (600,99.67) (650,99.67) (700,99.70) (800,99.70) (900,99.74) (950,99.75) (1050,99.78) (1200,99.79) (1300,99.82) (1450,99.84) (1600,99.85) (1750,99.89) (1950,99.92) (2150,99.94) (2350,99.93) (2600,99.96) (2850,99.97) (3150,99.97) (3450,99.98) (3850,99.98) (4200,99.99) (4650,99.99) (5100,99.99) (5650,99.99) (6200,100.00) (6850,100.00) (7550,100.00) (8350,100.00) (9200,100.00) (10100,100.00) (11150,100.00) (12250,100.00) (13500,100.00) (14900,100.00) (16400,100.00) (18100,100.00) (19900,100.00) (21950,100.00) (24200,100.00) (26650,100.00) (29350,100.00) (32350,100.00) (35600,100.00) (39250,100.00) (43200,100.00) (47600,100.00) (52450,100.00) (57800,100.00) (63650,100.00) (70150,100.00) (77250,100.00) (85100,100.00) (93750,100.00) (103300,100.00) (113750,100.00) (125350,100.00) (138050,100.00) (152100,100.00) (167550,100.00) (184550,100.00) (203300,100.00) (223950,100.00) (246700,100.00) (271800,100.00) (299400,100.00) (329800,100.00) (363300,100.00) (400250,100.00) (440900,100.00) (485650,100.00) (535000,100.00) (589350,100.00) (649250,100.00) (715200,100.00) (787850,100.00) (867850,100.00) (956050,100.00) (1053150,100.00) (1160150,100.00) (1278000,100.00) (1407800,100.00) (1550800,100.00) (1708350,100.00) (1881900,100.00) (2073050,100.00) (2283650,100.00) (2515650,100.00) (2771200,100.00) (3052700,100.00) (3362800,100.00) (3704400,100.00) (4080700,100.00) (4495200,100.00) (4951850,100.00) (5454850,100.00) (6008950,100.00) (6619400,100.00) (7291800,100.00) (8032500,100.00) (8848450,100.00) (9747300,100.00) (10737450,100.00) (11828200,100.00) (13029750,100.00) (14353300,100.00) (15811350,100.00)};
\addlegendentry{Estimated BATE prioritization}
\addplot[blue!70!black, solid] coordinates {(150,9.90) (200,10.41) (250,10.60) (300,12.19) (350,14.81) (400,14.06) (450,16.73) (500,14.69) (550,20.07) (600,17.62) (650,17.50) (700,20.87) (800,20.66) (900,21.53) (950,23.84) (1050,23.26) (1200,22.75) (1300,25.37) (1450,28.99) (1600,29.59) (1750,30.42) (1950,32.44) (2150,32.46) (2350,35.13) (2600,36.33) (2850,38.90) (3150,39.53) (3450,40.30) (3850,43.69) (4200,44.70) (4650,46.43) (5100,48.16) (5650,50.53) (6200,51.46) (6850,54.45) (7550,56.63) (8350,57.94) (9200,60.44) (10100,61.56) (11150,63.35) (12250,66.03) (13500,67.68) (14900,70.03) (16400,71.15) (18100,73.40) (19900,74.98) (21950,76.46) (24200,78.33) (26650,79.89) (29350,81.35) (32350,82.72) (35600,83.95) (39250,85.25) (43200,86.29) (47600,87.44) (52450,88.40) (57800,89.39) (63650,90.14) (70150,91.07) (77250,91.83) (85100,92.55) (93750,93.21) (103300,93.76) (113750,94.33) (125350,94.86) (138050,95.33) (152100,95.75) (167550,96.17) (184550,96.51) (203300,96.83) (223950,97.15) (246700,97.42) (271800,97.68) (299400,97.92) (329800,98.11) (363300,98.28) (400250,98.43) (440900,98.58) (485650,98.71) (535000,98.81) (589350,98.92) (649250,99.00) (715200,99.09) (787850,99.15) (867850,99.21) (956050,99.27) (1053150,99.34) (1160150,99.38) (1278000,99.44) (1407800,99.47) (1550800,99.51) (1708350,99.55) (1881900,99.59) (2073050,99.63) (2283650,99.67) (2515650,99.70) (2771200,99.73) (3052700,99.76) (3362800,99.78) (3704400,99.81) (4080700,99.83) (4495200,99.86) (4951850,99.87) (5454850,99.89) (6008950,99.91) (6619400,99.92) (7291800,99.93) (8032500,99.94) (8848450,99.95) (9747300,99.96) (10737450,99.97) (11828200,99.97) (13029750,99.97) (14353300,99.98) (15811350,99.98)};
\addlegendentry{CATE prioritization}
\addplot[green!55!black, densely dashdotted, domain=150:1.6e7, samples=2] {87.82};
\addlegendentry{Threshold prioritization}
\addplot[gray!60, dashdotted, domain=150:1.6e7, samples=2] {25.20};
\addlegendentry{Baseline prioritization}
\end{axis}
\end{tikzpicture}
\end{minipage}\hfill
\begin{minipage}[t]{0.49\textwidth}
\centering
\begin{tikzpicture}
\begin{axis}[
    width=\linewidth, height=6cm,
    title={\small (B) 75\% of effect variance forgone},
    xmode=log, xmin=150, xmax=1.6e7, ymin=0, ymax=102,
    xlabel={Experiment size $n$ (treated people)}, ylabel={Normalized Qini (\%)},
    xtick={1e3,1e4,1e5,1e6,1e7}, xticklabels={$10^3$,$10^4$,$10^5$,$10^6$,$10^7$},
    grid=major, grid style={gray!15},
    every axis plot/.append style={line width=1.3pt}, tick label style={font=\scriptsize}, label style={font=\small}
]
\addplot[orange!80!black, densely dashed] coordinates {(150,48.96) (200,49.05) (250,49.19) (300,49.26) (350,49.27) (400,49.35) (450,49.38) (500,49.38) (550,49.48) (600,49.46) (650,49.47) (700,49.49) (800,49.46) (900,49.52) (950,49.56) (1050,49.59) (1200,49.61) (1300,49.64) (1450,49.63) (1600,49.66) (1750,49.70) (1950,49.76) (2150,49.73) (2350,49.75) (2600,49.70) (2850,49.76) (3150,49.70) (3450,49.77) (3850,49.74) (4200,49.81) (4650,49.83) (5100,49.78) (5650,49.82) (6200,49.80) (6850,49.82) (7550,49.78) (8350,49.80) (9200,49.81) (10100,49.80) (11150,49.79) (12250,49.77) (13500,49.79) (14900,49.77) (16400,49.81) (18100,49.79) (19900,49.81) (21950,49.82) (24200,49.79) (26650,49.79) (29350,49.76) (32350,49.78) (35600,49.81) (39250,49.81) (43200,49.80) (47600,49.82) (52450,49.78) (57800,49.76) (63650,49.81) (70150,49.76) (77250,49.81) (85100,49.83) (93750,49.75) (103300,49.77) (113750,49.80) (125350,49.81) (138050,49.79) (152100,49.82) (167550,49.78) (184550,49.79) (203300,49.80) (223950,49.80) (246700,49.79) (271800,49.78) (299400,49.82) (329800,49.78) (363300,49.75) (400250,49.77) (440900,49.75) (485650,49.81) (535000,49.81) (589350,49.77) (649250,49.77) (715200,49.77) (787850,49.77) (867850,49.83) (956050,49.75) (1053150,49.81) (1160150,49.73) (1278000,49.81) (1407800,49.79) (1550800,49.82) (1708350,49.78) (1881900,49.79) (2073050,49.80) (2283650,49.83) (2515650,49.77) (2771200,49.77) (3052700,49.82) (3362800,49.83) (3704400,49.80) (4080700,49.77) (4495200,49.82) (4951850,49.82) (5454850,49.80) (6008950,49.77) (6619400,49.80) (7291800,49.81) (8032500,49.79) (8848450,49.81) (9747300,49.77) (10737450,49.79) (11828200,49.80) (13029750,49.81) (14353300,49.79) (15811350,49.81)};
\addplot[blue!70!black, solid] coordinates {(150,17.60) (200,19.93) (250,21.46) (300,24.06) (350,26.35) (400,27.42) (450,29.87) (500,30.08) (550,33.05) (600,33.37) (650,33.85) (700,36.22) (800,38.22) (900,39.74) (950,41.35) (1050,42.20) (1200,44.20) (1300,46.35) (1450,49.50) (1600,50.58) (1750,52.12) (1950,54.72) (2150,55.89) (2350,58.43) (2600,60.13) (2850,62.28) (3150,63.83) (3450,65.94) (3850,67.60) (4200,69.38) (4650,71.15) (5100,73.05) (5650,74.76) (6200,76.17) (6850,77.88) (7550,79.27) (8350,80.77) (9200,82.26) (10100,83.42) (11150,84.61) (12250,85.80) (13500,86.89) (14900,88.05) (16400,88.82) (18100,89.79) (19900,90.64) (21950,91.42) (24200,92.16) (26650,92.75) (29350,93.43) (32350,93.96) (35600,94.45) (39250,94.92) (43200,95.40) (47600,95.79) (52450,96.19) (57800,96.53) (63650,96.83) (70150,97.13) (77250,97.40) (85100,97.63) (93750,97.86) (103300,98.06) (113750,98.21) (125350,98.40) (138050,98.54) (152100,98.68) (167550,98.81) (184550,98.91) (203300,99.02) (223950,99.10) (246700,99.18) (271800,99.26) (299400,99.33) (329800,99.39) (363300,99.45) (400250,99.49) (440900,99.55) (485650,99.58) (535000,99.62) (589350,99.66) (649250,99.69) (715200,99.71) (787850,99.74) (867850,99.76) (956050,99.78) (1053150,99.80) (1160150,99.82) (1278000,99.83) (1407800,99.85) (1550800,99.86) (1708350,99.87) (1881900,99.88) (2073050,99.89) (2283650,99.90) (2515650,99.91) (2771200,99.92) (3052700,99.93) (3362800,99.93) (3704400,99.94) (4080700,99.94) (4495200,99.95) (4951850,99.95) (5454850,99.96) (6008950,99.96) (6619400,99.96) (7291800,99.97) (8032500,99.97) (8848450,99.97) (9747300,99.98) (10737450,99.98) (11828200,99.98) (13029750,99.98) (14353300,99.99) (15811350,99.99)};
\addplot[green!55!black, densely dashdotted, domain=150:1.6e7, samples=2] {41.98};
\addplot[gray!60, dashdotted, domain=150:1.6e7, samples=2] {9.46};
\end{axis}
\end{tikzpicture}
\end{minipage}
\par\medskip
\pgfplotslegendfromname{straddleleg}
\caption{\textbf{Theory gets close, a little data fixes the order, and recovering forgone heterogeneity takes more.} Normalized Qini in the straddling population of the running example (\cref{sec:example}), by experiment size. Estimated BATE prioritization uses the one-parameter logit. (A) A common shift. (B) Shifts that also vary within baseline groups ($r=0.4$).}
\label{fig:straddle}
\end{figure}

As the shift grows, however, the true peak moves farther below 50\%, and baseline prioritization can outperform threshold prioritization. For example, with a shift of 2 rather than 0.5 in the straddling population, ranking from the lowest baseline up has a normalized Qini of 78\%, while threshold prioritization does slightly worse than random targeting. The baseline alone cannot tell us when this happens or which ranking is better. For that, we need experimental data.

Learning the BATE replaces the guess with an estimate from an experiment. Under a common shift, the BATE equals the CATE, and \cref{prop:onenumber} shows that the optimal ranking can be recovered by estimating only the shift. With logistic noise, this means estimating a single treatment coefficient. An estimate $\hat\delta$ misplaces the peak by $|\hat\delta-\delta|/2$, but \cref{prop:secondorder} bounds the resulting loss by $\bar gL_\delta(\hat\delta-\delta)^2/4$. An estimator with mean squared error of order $1/n$ therefore incurs expected loss of at most order $1/n$.

\Cref{fig:straddle}A illustrates this in the straddling population. Without any experimental data, baseline prioritization has a normalized Qini of 25\% when ranking from the highest baseline down, and threshold prioritization reaches 88\%. Estimated BATE prioritization, using the one-parameter logit (\cref{sec:example}), ranks people by their distance from the estimated peak. With just 150 treated people, it reaches 99\%. CATE prioritization needs about 650{,}000 treated people to reach that level with the cell estimator.

The one-parameter result depends on the common shift and the known link. If the shift varies with the baseline, or the link is asymmetric, as with a complementary log-log link, learning the BATE still involves a function of one variable, and BATE prioritization remains the best use of baseline information. But when effects vary among people with the same baseline, an experiment has more to discover. In \cref{fig:straddle}B, shifts vary within baseline groups, leaving 75\% of effect variance unexplained by the baseline. The estimated BATE prioritization now attains a normalized Qini of only 50\% with 150 treated people. CATE prioritization overtakes it at about 1{,}600 treated people and keeps improving. The next section examines when recovering this forgone heterogeneity is worth the additional experimental data.

% =====================================================================
\section{Learning What the Baseline Misses}\label{sec:size}
% =====================================================================

Once the baseline is put in the right order, further gains require information it does not contain. Throughout this section, we assume the BATE is monotone in the baseline, so baseline prioritization ranks people as BATE prioritization does; the question is no longer how to order the baseline but what it cannot see. People with the same baseline may have different effects; their average within-group variance is the \textbf{forgone heterogeneity}, $\mathcal U=\E[\Var(\beta\mid\theta)]$. The rest of effect variation is visible through the BATE, so the baseline's potential is
\begin{equation}\label{eq:potential}
  R^2_{\beta\mid\theta}=\frac{\Var(m_\theta)}{\Var(m_\theta)+\mathcal U}.
\end{equation}
We call $1-R^2_{\beta\mid\theta}=\mathcal U/\Var(\beta)$ the \textbf{forgone share}.
In the threshold model, the baseline reveals where a person's systematic index sits relative to the threshold and may also reveal something about how much treatment shifts that index. We therefore split the shift into the part the baseline can predict and a part $\eta$ it cannot:
\begin{equation}
  \delta(X)=\E[\delta\mid\mu(X)]+\eta,\qquad \E[\eta\mid\mu]=0 .
\end{equation}
Since $\theta$ and $\mu$ carry the same information, people who share a baseline differ in their effects only through $\eta$. How much does the baseline miss as a result? We formalize an answer to this question in the next proposition.

\begin{proposition}[What the baseline misses]\label{prop:unseen}
Under \cref{ass:threshold}:
\begin{enumerate}[label=(\alph*)]
  \item If the shift is predictable from the systematic index, $\eta=0$, then $\mathcal U=0$, however much $\delta$ varies.
  \item Let $\eta=tZ$ with $\E[Z\mid\mu]=0$ and $\E[Z^4]<\infty$, and suppose $f'$ is bounded. As $t\to0$,
  \begin{equation}\label{eq:unseen}
    \mathcal U=\E\big[\underbrace{f(\mu+\E[\delta\mid\mu])^2}_{\text{sensitivity}}\,\underbrace{\Var(\eta\mid\mu)}_{\text{unpredictable shifts}}\big]+O(t^3).
  \end{equation}
  \end{enumerate}
\end{proposition}

A small unpredictable shift $\eta$ changes the treated probability by about $f(\mu+\E[\delta\mid\mu])\,\eta$. The density $f$ measures sensitivity at the expected treated index. The baseline therefore misses much only where the same people are both sensitive to treatment and subject to shifts it cannot predict. Either ingredient alone is insufficient.

Shifts can vary a great deal and still be captured by the baseline. For instance, suppose subscribers differ only in how much they care about price: caring more lowers the chance of renewal and raises the response to a discount. The baseline then reveals the shift, so its potential stays at one. Total effect variation alone thus says little about what CATE prioritization can add. Shifts can also vary among people with the same baseline and still matter little. A rebate that makes purchase nearly certain for everyone gives each person an effect close to $1-\theta$, however differently they value it. In the swimming-pool analogy, differences in the strength of the push matter most for people the typical push would leave right at the edge, where a slightly harder or softer push decides whether they fall in. They matter little for people the push sends in easily, or leaves far from the water.

\subsection{What Determines the Experiment Size Needed?}\label{sec:thebar}

An effect estimate gains the forgone heterogeneity but adds estimation noise, and the noise shrinks as the experiment grows. CATE prioritization therefore wins once the experiment is large enough for the gain to outweigh the noise. To reason about the size needed, we approximate effects, baselines, and their estimation errors as jointly Gaussian (\cref{app:gaussian}).

How fast the estimation noise shrinks with more data depends on the estimator. With $K$ equally likely cells and $n/K$ treated people in each, the cell estimator's noise variance is about $K\bar w/n$, where $\bar w=\E[\theta_1(1-\theta_1)]$ is the average outcome variance under treatment. A finer partition can capture more of the heterogeneity in effects, but more cells, like noisier outcomes, mean more noise at a given experiment size. The estimate's signal-to-noise ratio, total effect variance over noise variance, is then $n\Var(\beta)/(K\bar w)$, which grows in proportion to the experiment. 

Under the Gaussian approximation, CATE prioritization uniformly dominates baseline prioritization, and so has a higher normalized Qini, if and only if the experiment has
\begin{equation}\label{eq:nstar}
  n\;\ge\;n^*=\frac{K\bar w}{\Var(\beta)}\cdot\frac{R^2_{\beta\mid\theta}}{1-R^2_{\beta\mid\theta}}
\end{equation}
treated people. The first factor is the experiment needed to bring the signal-to-noise ratio to one. It reflects the estimator and how much effects vary relative to noise. The second factor reflects how much effect heterogeneity the baseline captures relative to what it misses. For example, with treated outcome probabilities near 10\%, 50 cells, and effects with a standard deviation of 2 percentage points, the first factor is about 11{,}250 treated people. The second is 1, 4, and 19 when half, a fifth, and a twentieth of effect variance is forgone, so $n^*$ is about 11{,}250, 45{,}000, and 213{,}750. Total effect variation is the same in all three cases, yet the experiment needed varies nearly twentyfold with how much of it the baseline misses.

\Cref{tab:avg} shows how the experiment needed changes in the running example, where the Gaussian assumption does not hold. The approximation aligns well with the simulation results. As the share of effect variance the baseline misses falls from over 80\% to under 10\%, the size at which CATE prioritization wins in normalized Qini rises from 550 to more than 100{,}000 treated people, and $n^*$ stays within about 25\% of it. 

Treating the baseline as known might seem to stack the comparison against effect estimation, since in practice the baseline is estimated too. The opposite holds in most scenarios: estimating the baseline from the experiment roughly doubles the cell estimator's noise but barely affects baseline prioritization (\cref{sec:easier}), so it raises the size CATE prioritization needs to win by 20\% to 140\% in seven of the eight scenarios (\cref{app:learned}).

\begin{table}[t]
\centering
\caption{\textbf{When CATE prioritization wins in normalized Qini.} The crossover is the smallest experiment size (number of treated people) at which CATE prioritization consistently matches or exceeds baseline prioritization in normalized Qini. Observed crossovers come from simulations of the running example; predicted ones from the Gaussian formula $n^*$ in \Cref{eq:nstar}. The forgone share is the share of effect variance the baseline misses, and $r$ is the standard deviation of unpredictable shifts relative to the expected shift.}
\label{tab:avg}
\vspace{4pt}
\small
\begin{tabular}{lcrrr}
\toprule
& & & \multicolumn{2}{c}{Crossover} \\
\cmidrule(l){4-5}
Population & $r$ & Forgone share & Predicted, $n^*$ & Observed \\
\midrule
Rare   & 0.8 & 81\% & 420 & 550 \\
Rare   & 0.4 & 53\% & 4{,}100 & 5{,}100 \\
Rare   & 0.2 & 22\% & 27{,}500 & 32{,}350 \\
Rare   & 0.1 & 7\% & 132{,}100 & 125{,}350 \\
Common & 0.8 & 72\% & 640 & 700 \\
Common & 0.4 & 37\% & 5{,}800 & 7{,}550 \\
Common & 0.2 & 13\% & 32{,}200 & 35{,}600 \\
Common & 0.1 & 4\% & 142{,}600 & 113{,}750 \\
\bottomrule
\end{tabular}
\end{table}

\subsection{The Role of Treatment Capacity}\label{sec:narrow}

When we have the capacity to treat only a fraction $u$ of people, a ranking matters only through whom it places above or below the cutoff. That means both estimation errors and forgone heterogeneity can change the allocation only if they move someone across that cutoff. Whether CATE prioritization improves the allocation therefore depends on what happens at the margin, and the margin moves with capacity.

To make this precise, we compare CATE prioritization with baseline prioritization at a fixed treatment fraction $u$. Under this section's monotonicity assumption, ranking people by their baselines is equivalent to ranking them by their BATEs. We can therefore express the treatment cutoff on the BATE scale. Let $c_u$ satisfy $\mathbb P(m_\theta\ge c_u)=u$.

To see what we can gain or lose by changing who gets treated, consider a CATE estimate that moves someone above the cutoff, displacing someone else. The swap helps if the newcomer's true effect is larger and hurts if it is smaller, as when estimation error rather than a real difference in effects moved them up. We therefore distinguish the true departure from the BATE, $\beta-m_\theta=t\zeta$, from the estimated departure, $\hat\beta-m_\theta=t\xi$. The parameter $t>0$ scales both, so small departures change treatment only for people near the cutoff. By definition, $\E[\zeta\mid\theta]=0$. We also assume that estimates average to the BATE among people with the same baseline, $\E[\hat\beta\mid\theta]=m_\theta$. We treat the fitted model as fixed and take expectations over the population.

\begin{samepage}
\begin{proposition}[The trade-off at the cutoff]\label{thm:margin}
Under regularity conditions given in \cref{app:proofs5}, as $t\to0$,
\begin{equation}
  V(u;\hat\beta)-V(u;m_\theta)=\frac{g_{m_\theta}(c_u)}{2}\Big\{\underbrace{\Var(\beta\mid m_\theta=c_u)}_{\text{forgone at the margin}}-\underbrace{\E\big[(\hat\beta-\beta)^2\mid m_\theta=c_u\big]}_{\text{error at the margin}}\Big\}+o(t^2),
\end{equation}
where $g_{m_\theta}$ is the density of the BATE.
\end{proposition}
\end{samepage}

The proposition implies that the people close to the cutoff are the ones that matter. Their forgone heterogeneity is what an estimate can gain, and its error among them is what it loses, so CATE prioritization beats baseline prioritization at fraction $u$ when the first exceeds the second. The error is a mean squared error, so it counts bias as well as noise. How many people sit at the cutoff, $g_{m_\theta}(c_u)$, scales gain and loss alike, so it affects how much is at stake but not which ranking wins. Which capacities favor CATE prioritization depends on how forgone heterogeneity and error change along the ranking. The result is local: larger departures widen the band of people who can change sides, and the same trade-off then plays out over that band.

Let's consider the error first. For an unbiased estimate, the error at the margin is just estimation noise, and that noise shrinks as the experiment grows. We can therefore ask how large an experiment must be for CATE prioritization to win at a given capacity, the counterpart at the margin of $n^*$ in \cref{sec:thebar}. The answer depends on how much of the estimate's variance among people at the cutoff is signal rather than noise. For the cell estimator, each person's estimate has noise variance $K\,w(X)/n$, where $w=\theta_1(1-\theta_1)$ is the outcome variance under treatment. Treating the cells as a continuous population, CATE prioritization wins at fraction $u$ once
\begin{equation}\label{eq:margin}
  n\;\ge\;n_{\mathrm{margin}}^*(u)=\frac{K\,\E\big[w(X)\mid m_\theta=c_u\big]}{\mathcal U_u},
\end{equation}
where $\mathcal U_u=\Var(\beta\mid m_\theta=c_u)$ is the forgone heterogeneity among people at the cutoff. The numerator is outcome noise among them. So, CATE prioritization wins at fraction $u$ once more than half of the estimate's variance among people at the cutoff is signal.

Outcome noise is largest for people whose treated probability is near 50\%. So is forgone heterogeneity, and it grows faster: it scales with the square of sensitivity (\cref{prop:unseen}), while under logistic noise outcome noise equals sensitivity itself, $F(1-F)=f$, at the treated index. When unpredictable shifts are equally variable, the most sensitive people therefore offer the most to learn per unit of noise. These are typically also the people baseline prioritization ranks first, those closest to the effect peak, so a narrow campaign places the margin where learning is cheapest.

\begin{corollary}[Narrow campaigns need less]\label{cor:narrow}
Suppose the noise is logistic, everyone has the same expected shift $\E[\delta\mid\mu]=\bar\delta>0$ and the same variance of unpredictable shifts $\Var(\eta\mid\mu)=v>0$, and the population lies on one side of the effect peak $\mu=-\bar\delta/2$, away from it. On the rising side, suppose also that everyone's probability under the expected shift, $F(\mu+\bar\delta)$, is at most $1/2$. Then, to leading order in small unpredictable shifts, $n_{\mathrm{margin}}^*(u)$ in \eqref{eq:margin} is nondecreasing in $u$.
\end{corollary}

The condition on the rising side keeps treated probabilities below 50\%, so that sensitivity rises up the ranking. \Cref{tab:fraction} shows the pattern holds in the running example, even with unpredictable shifts nearly as large as the expected shift ($r=0.8$). With 53\% forgone heterogeneity (second row), CATE prioritization wins among the top 10\% with 1{,}750 treated people but needs 12{,}250 among the top 80\%.

Because the experiment needed rises with capacity, a win in normalized Qini need not extend to broad campaigns. With 5{,}100 treated people, enough to win in normalized Qini (\cref{tab:avg}, second row), CATE prioritization still loses among the top half and the top 80\%, and winning at every capacity, where it happens at all, takes roughly two to six times $n^*$ (\cref{tab:fraction}, last column). In a continuous population, the noise level at which the two rankings tie is within 13\% of the prediction of \cref{thm:margin} at fractions from 20\% to 80\% (\cref{app:simdetails}).

\begin{table}[t]
\centering
\caption{\textbf{Narrow campaigns usually need less experimental data, but not always.} Smallest experiment size (number of treated people) at which CATE prioritization consistently matches or exceeds the total effect achieved by baseline prioritization when treating the top 10\%, 50\%, or 80\% of the ranking, or at every treatment fraction (``Everywhere''), in simulations of the running example. $n^*$ is the normalized-Qini prediction from \cref{tab:avg}, shown for reference; other columns as in \cref{tab:avg}. A dash means no win up to 15 million. The last row is a variant of the rare population in which unpredictable shifts are largest at the bottom of the ranking.}
\label{tab:fraction}
\vspace{4pt}
\small
\begin{tabular}{lcrrrrrr}
\toprule
Population & $r$ & Forgone share & $n^*$ & Top 10\% & Top 50\% & Top 80\% & Everywhere \\
\midrule
Rare   & 0.8 & 81\% & 420 & 200 & 650 & 900 & 950 \\
Rare   & 0.4 & 53\% & 4{,}100 & 1{,}750 & 6{,}850 & 12{,}250 & 13{,}500 \\
Rare   & 0.2 & 22\% & 27{,}500 & 11{,}150 & 47{,}600 & 77{,}250 & 77{,}250 \\
Rare   & 0.1 & 7\% & 132{,}100 & 70{,}150 & 223{,}950 & 440{,}900 & 440{,}900 \\
Common & 0.8 & 72\% & 640 & 150 & 900 & 1{,}200 & 1{,}200 \\
Common & 0.4 & 37\% & 5{,}800 & 1{,}300 & 11{,}150 & 32{,}350 & 32{,}350 \\
Common & 0.2 & 13\% & 32{,}200 & 10{,}100 & 70{,}150 & 152{,}100 & 167{,}550 \\
Common & 0.1 & 4\% & 142{,}600 & 70{,}150 & 715{,}200 & -- & -- \\
\midrule
Reversed &  & 38\% & 11{,}900 & 43{,}200 & 13{,}500 & 4{,}650 & 52{,}450 \\
\bottomrule
\end{tabular}
\end{table}

This pattern assumes the unpredictable shifts vary about equally across baselines, as in the simulations. If shifts are much larger far from the effect peak, the pattern can reverse. In a reversed variant of the rare population (last row), the spread of unpredictable shifts is about nine times larger in the lowest baseline group than in the highest, enough to outweigh the lower sensitivity there: winning among the top 10\% takes 43{,}200 treated people, but among the top 80\% only 4{,}650.

The value of experimental data for treatment prioritization should therefore be assessed at the capacity at which treatment decisions will be made.

% =====================================================================
\section{Evidence from an Advertising Experiment}\label{sec:criteo}
% =====================================================================

We illustrate these results with the Criteo uplift data \citep{diemert2018large, diemert2021benchmark}, a randomized advertising experiment with about 14 million users, 85\% of whom were assigned to receive ads. The outcome is whether a user visits the advertiser's site, which 3.8\% of control users and 4.9\% of treated users do.

We mimic an organization that already holds untreated data and runs an experiment of growing size. A random 30\% of users from both arms is held out for evaluation. The remaining 1.5 million control users train a gradient-boosted baseline classifier on the 12 anonymized features; this model stays fixed. An experiment of size $n$ adds $n$ treated users drawn from the remaining 8.3 million, so sizes count treated users only. Because the untreated pool is already large, more experimental controls would add little, so we vary only the treated sample. On held-out data, the prioritization curve at fraction $u$ is estimated as $u$ times the difference between treated and control visit rates among the top $u$, reported as incremental visits per 1{,}000 users (\cref{app:criteo}). This experiment lets us examine both tasks: fixing the order and learning what the baseline misses.

Most users are unlikely to visit without ads. The median baseline is 0.3\%, and only 1.8\% of users have a baseline above 50\%. \Cref{fig:criteo}A shows that effects rise with the baseline, peak between 40\% and 50\%, and fall to nearly zero above 70\%. A common shift in the log-odds, fitted to treated users, implies a peak at 47\% and tracks the rise closely, although effects fall faster above the peak. The population therefore straddles the effect peak, but barely, with nearly everyone on the rising side.

This is the case in which baseline prioritization should work well except among the users ranked highest, and threshold prioritization should repair the top (\cref{sec:enough}). \Cref{tab:criteo} shows both. Among the top 1\%, ranking from the highest baseline down yields 0.29 incremental visits per 1{,}000 users, because it starts with users who would visit anyway. Threshold prioritization doubles that to 0.61 without any experimental data. From the top 10\% on, the two rankings select almost the same users. Estimated BATE prioritization with the one-parameter logit matches threshold prioritization from 1{,}000 treated users on. The fitted shift, about 0.26, is modest, so ranking by closeness to 50\% is already close to the best use of the baseline.

The comparison with CATE prioritization is informative only if the effect estimates are competitive. We use three learners, all based on gradient boosting. The T-learner \citep{kunzel2019metalearners} fits a model of treated outcomes and subtracts the baseline; learners of this kind remain competitive with recent deep uplift models on these data \citep{liu2024benchmarking}. The offset learner uses the baseline log-odds as an offset, so it only learns how treated outcomes depart from the baseline. The X-learner \citep{kunzel2019metalearners}, designed for experiments with very unequal arms, as here, also learns from control users.

\begin{table}[t]
\centering
\caption{\textbf{Value $V(u;\cdot)$ of each ranking in the Criteo experiment.} Incremental visits per 1{,}000 users from treating the top 1\% or the top 50\%, by experiment size in thousands of treated users, averaged over repeated experiments. The largest size, 8{,}316 thousand, is the whole pool of treated users. Baseline and threshold prioritization use no experimental data.}
\label{tab:criteo}
\vspace{4pt}
\small
\begin{tabular}{lrrrrrrrrrr}
\toprule
& \multicolumn{5}{c}{Top 1\%} & \multicolumn{5}{c}{Top 50\%} \\
\cmidrule(lr){2-6}\cmidrule(l){7-11}
Experiment size (thousands) &  1 & 10 & 100 & 1{,}000 & 8{,}316 & 1 & 10 & 100 & 1{,}000 & 8{,}316 \\
\midrule
Baseline & \multicolumn{5}{c}{0.29} & \multicolumn{5}{c}{10.23} \\
Threshold & \multicolumn{5}{c}{0.61} & \multicolumn{5}{c}{10.23} \\
BATE, one-parameter logit & 0.62 & 0.62 & 0.62 & 0.63 & 0.64 & 10.23 & 10.23 & 10.23 & 10.23 & 10.23 \\
CATE, T-learner & 0.83 & 1.23 & 1.37 & 1.60 & 1.71 & 5.97 & 7.86 & 8.70 & 9.77 & 9.88 \\
CATE, offset learner & 0.83 & 1.20 & 1.50 & 1.73 & 1.69 & 7.28 & 8.52 & 9.44 & 9.99 & 9.88 \\
CATE, X-learner & 0.66 & 1.31 & 1.53 & 1.73 & 1.83 & 8.48 & 8.95 & 9.44 & 10.14 & 10.19 \\
\bottomrule
\end{tabular}
\end{table}

Learning beyond the baseline pays at some capacities long before others. Among the top 1\%, all three learners beat estimated BATE prioritization with a thousand treated users. With a million, they nearly triple its value, from 0.63 to about 1.7 incremental visits per 1{,}000 users: the average effect among those treated rises from 6 to 17 percentage points. The X-learner, the best of the three, wins among the top 10\% with 3{,}000 treated users and among the top 25\% with 300{,}000 (\cref{fig:criteo}B); the other two CATE learners need about 10{,}000 and a million.

Among the top half, even all 8.3 million treated users bring no detectable gain for the X-learner (10.19 against 10.23 incremental visits per 1{,}000 users), and the other two fall short (9.88). \Cref{fig:criteo}B shows the X-learner's full profile: small experiments gain at narrow fractions but lose at broad ones. The same experiment that triples the value of a narrow campaign brings no detectable gain to a broad one. This pattern is consistent with the cutoff mechanism in \cref{sec:narrow}. Under the conditions of \cref{cor:narrow}, a narrow campaign places the cutoff among users whose outcomes are sensitive to ads, where unpredictable shifts create more effect variation relative to outcome noise. A broad campaign places it among users who almost never visit, where differences in shifts barely change their effects. This offers a mechanism for the observed pattern, although we cannot observe forgone heterogeneity directly.

\begin{figure}[t]
\centering
\begin{minipage}[t]{0.49\textwidth}
\centering
\begin{tikzpicture}
\begin{axis}[
    width=\linewidth, height=6cm,
    title={\small (A) Effects by baseline},
    xmode=log, xmin=4e-4, xmax=1, ymin=0, ymax=9,
    xlabel={Baseline}, ylabel={Effect (percentage points)},
    xtick={0.001,0.01,0.1,1}, xticklabels={0.1\%,1\%,10\%,100\%},
    grid=major, grid style={gray!15},
    tick label style={font=\scriptsize}, label style={font=\small}
]
\addplot[orange!80!black, densely dashed, line width=1.3pt] coordinates {(0.0005,0.01) (0.0006,0.02) (0.0007,0.02) (0.0008,0.02) (0.0009,0.03) (0.0011,0.03) (0.0012,0.04) (0.0014,0.04) (0.0016,0.05) (0.0018,0.05) (0.0020,0.06) (0.0023,0.07) (0.0026,0.08) (0.0029,0.09) (0.0033,0.10) (0.0038,0.11) (0.0043,0.12) (0.0048,0.14) (0.0055,0.16) (0.0062,0.18) (0.0071,0.21) (0.0080,0.23) (0.0091,0.26) (0.0103,0.30) (0.0117,0.34) (0.0133,0.38) (0.0150,0.43) (0.0171,0.49) (0.0194,0.55) (0.0220,0.63) (0.0249,0.71) (0.0283,0.80) (0.0321,0.90) (0.0364,1.02) (0.0412,1.15) (0.0468,1.29) (0.0531,1.45) (0.0602,1.63) (0.0683,1.83) (0.0775,2.05) (0.0879,2.29) (0.0997,2.56) (0.1131,2.85) (0.1283,3.16) (0.1455,3.50) (0.1651,3.86) (0.1872,4.23) (0.2124,4.62) (0.2409,5.01) (0.2733,5.39) (0.3100,5.75) (0.3517,6.06) (0.3989,6.30) (0.4525,6.41) (0.5133,6.37) (0.5823,6.09) (0.6606,5.51) (0.7493,4.52) (0.8500,2.99)};
\addplot[black, only marks, mark=*, mark size=1.6pt, error bars/.cd, y dir=both, y explicit, error bar style={line width=0.6pt}] coordinates {(0.0008,0.01) +- (0,0.01) (0.0013,0.05) +- (0,0.02) (0.0036,0.11) +- (0,0.04) (0.0069,0.20) +- (0,0.07) (0.0141,0.40) +- (0,0.11) (0.0315,1.23) +- (0,0.17) (0.0716,2.46) +- (0,0.35) (0.1421,3.72) +- (0,0.52) (0.2471,5.27) +- (0,0.81) (0.3491,5.98) +- (0,1.06) (0.4479,6.61) +- (0,1.16) (0.5458,6.09) +- (0,1.41) (0.6456,4.30) +- (0,1.76) (0.7443,0.84) +- (0,2.35)};
\node[font=\scriptsize, text=orange!80!black, anchor=west, align=left] at (axis cs:0.0012,2.4) {common shift\\in the log-odds};
\draw[black, dotted, line width=0.8pt] (axis cs:0.5,0) -- (axis cs:0.5,9);
\end{axis}
\end{tikzpicture}
\end{minipage}\hfill
\begin{minipage}[t]{0.49\textwidth}
\centering
\begin{tikzpicture}
\begin{axis}[
    width=\linewidth, height=6cm,
    title={\small (B) Gain from CATE prioritization},
    xmode=log, xmin=0.01, xmax=1, ymin=-1.4, ymax=2,
    xlabel={Treatment fraction $u$}, ylabel={Gain over BATE (visits per 1{,}000)},
    xtick={0.01,0.02,0.05,0.1,0.2,0.5,1}, xticklabels={1\%,2\%,5\%,10\%,20\%,50\%,100\%},
    legend style={font=\scriptsize, draw=none, fill=white, fill opacity=0.85, text opacity=1, at={(0.03,0.03)}, anchor=south west},
    legend cell align=left, grid=major, grid style={gray!15},
    every axis plot/.append style={line width=1.2pt}, tick label style={font=\scriptsize}, label style={font=\small}
]
\addplot[black!60, line width=0.6pt, domain=0.01:1, samples=2, forget plot] {0};
\addplot[blue!25!white] coordinates {(0.01,0.530) (0.02,0.601) (0.03,0.799) (0.04,0.780) (0.05,0.680) (0.06,0.575) (0.07,0.408) (0.08,0.232) (0.09,0.199) (0.10,0.121) (0.11,0.105) (0.12,0.047) (0.13,-0.103) (0.14,-0.134) (0.15,-0.248) (0.16,-0.347) (0.17,-0.399) (0.18,-0.387) (0.19,-0.437) (0.20,-0.513) (0.21,-0.509) (0.22,-0.497) (0.23,-0.498) (0.24,-0.571) (0.25,-0.643) (0.26,-0.698) (0.27,-0.748) (0.28,-0.824) (0.29,-0.813) (0.30,-0.835) (0.31,-0.889) (0.32,-0.933) (0.33,-0.971) (0.34,-0.989) (0.35,-1.038) (0.36,-1.057) (0.37,-1.074) (0.38,-1.081) (0.39,-1.103) (0.40,-1.110) (0.41,-1.124) (0.42,-1.140) (0.43,-1.145) (0.44,-1.138) (0.45,-1.148) (0.46,-1.142) (0.47,-1.176) (0.48,-1.179) (0.49,-1.199) (0.50,-1.203) (0.51,-1.194) (0.52,-1.201) (0.53,-1.201) (0.54,-1.191) (0.55,-1.185) (0.56,-1.189) (0.57,-1.160) (0.58,-1.163) (0.59,-1.166) (0.60,-1.174) (0.61,-1.173) (0.62,-1.161) (0.63,-1.174) (0.64,-1.176) (0.65,-1.184) (0.66,-1.187) (0.67,-1.196) (0.68,-1.197) (0.69,-1.188) (0.70,-1.172) (0.71,-1.174) (0.72,-1.168) (0.73,-1.156) (0.74,-1.144) (0.75,-1.128) (0.76,-1.110) (0.77,-1.067) (0.78,-1.043) (0.79,-1.018) (0.80,-0.990) (0.81,-0.959) (0.82,-0.941) (0.83,-0.933) (0.84,-0.919) (0.85,-0.910) (0.86,-0.894) (0.87,-0.876) (0.88,-0.843) (0.89,-0.793) (0.90,-0.727) (0.91,-0.672) (0.92,-0.600) (0.93,-0.530) (0.94,-0.477) (0.95,-0.430) (0.96,-0.388) (0.97,-0.344) (0.98,-0.272) (0.99,-0.173) (1.00,0.000)};
\addlegendentry{3{,}000}
\addplot[blue!45!white] coordinates {(0.01,0.742) (0.02,0.917) (0.03,1.085) (0.04,1.010) (0.05,0.978) (0.06,0.916) (0.07,0.800) (0.08,0.613) (0.09,0.534) (0.10,0.481) (0.11,0.415) (0.12,0.364) (0.13,0.210) (0.14,0.184) (0.15,0.081) (0.16,-0.029) (0.17,-0.089) (0.18,-0.082) (0.19,-0.134) (0.20,-0.233) (0.21,-0.239) (0.22,-0.248) (0.23,-0.260) (0.24,-0.323) (0.25,-0.393) (0.26,-0.449) (0.27,-0.508) (0.28,-0.567) (0.29,-0.553) (0.30,-0.570) (0.31,-0.633) (0.32,-0.686) (0.33,-0.723) (0.34,-0.735) (0.35,-0.786) (0.36,-0.797) (0.37,-0.809) (0.38,-0.827) (0.39,-0.851) (0.40,-0.847) (0.41,-0.853) (0.42,-0.877) (0.43,-0.884) (0.44,-0.888) (0.45,-0.905) (0.46,-0.906) (0.47,-0.945) (0.48,-0.953) (0.49,-0.976) (0.50,-0.985) (0.51,-0.982) (0.52,-0.996) (0.53,-1.001) (0.54,-0.994) (0.55,-0.993) (0.56,-1.003) (0.57,-0.982) (0.58,-0.983) (0.59,-0.991) (0.60,-0.997) (0.61,-1.002) (0.62,-0.993) (0.63,-1.009) (0.64,-1.017) (0.65,-1.040) (0.66,-1.059) (0.67,-1.073) (0.68,-1.075) (0.69,-1.071) (0.70,-1.060) (0.71,-1.067) (0.72,-1.058) (0.73,-1.047) (0.74,-1.039) (0.75,-1.023) (0.76,-1.009) (0.77,-0.997) (0.78,-0.978) (0.79,-0.958) (0.80,-0.935) (0.81,-0.907) (0.82,-0.882) (0.83,-0.861) (0.84,-0.849) (0.85,-0.829) (0.86,-0.807) (0.87,-0.787) (0.88,-0.761) (0.89,-0.733) (0.90,-0.710) (0.91,-0.672) (0.92,-0.630) (0.93,-0.578) (0.94,-0.523) (0.95,-0.472) (0.96,-0.412) (0.97,-0.341) (0.98,-0.250) (0.99,-0.128) (1.00,0.000)};
\addlegendentry{30{,}000}
\addplot[blue!70!black] coordinates {(0.01,1.015) (0.02,1.233) (0.03,1.450) (0.04,1.443) (0.05,1.393) (0.06,1.361) (0.07,1.249) (0.08,1.043) (0.09,0.948) (0.10,0.906) (0.11,0.808) (0.12,0.746) (0.13,0.585) (0.14,0.558) (0.15,0.477) (0.16,0.394) (0.17,0.336) (0.18,0.329) (0.19,0.276) (0.20,0.213) (0.21,0.203) (0.22,0.214) (0.23,0.224) (0.24,0.183) (0.25,0.143) (0.26,0.091) (0.27,0.049) (0.28,0.007) (0.29,0.025) (0.30,0.016) (0.31,-0.026) (0.32,-0.062) (0.33,-0.093) (0.34,-0.094) (0.35,-0.140) (0.36,-0.142) (0.37,-0.134) (0.38,-0.137) (0.39,-0.160) (0.40,-0.151) (0.41,-0.144) (0.42,-0.147) (0.43,-0.141) (0.44,-0.138) (0.45,-0.154) (0.46,-0.148) (0.47,-0.170) (0.48,-0.167) (0.49,-0.181) (0.50,-0.178) (0.51,-0.155) (0.52,-0.163) (0.53,-0.153) (0.54,-0.141) (0.55,-0.137) (0.56,-0.146) (0.57,-0.121) (0.58,-0.130) (0.59,-0.142) (0.60,-0.146) (0.61,-0.146) (0.62,-0.137) (0.63,-0.160) (0.64,-0.167) (0.65,-0.185) (0.66,-0.210) (0.67,-0.229) (0.68,-0.236) (0.69,-0.235) (0.70,-0.230) (0.71,-0.243) (0.72,-0.241) (0.73,-0.244) (0.74,-0.242) (0.75,-0.238) (0.76,-0.234) (0.77,-0.226) (0.78,-0.222) (0.79,-0.216) (0.80,-0.213) (0.81,-0.202) (0.82,-0.195) (0.83,-0.197) (0.84,-0.196) (0.85,-0.195) (0.86,-0.192) (0.87,-0.192) (0.88,-0.198) (0.89,-0.196) (0.90,-0.198) (0.91,-0.195) (0.92,-0.197) (0.93,-0.191) (0.94,-0.185) (0.95,-0.186) (0.96,-0.181) (0.97,-0.162) (0.98,-0.137) (0.99,-0.058) (1.00,0.000)};
\addlegendentry{300{,}000}
\addplot[blue!35!black] coordinates {(0.01,1.196) (0.02,1.465) (0.03,1.651) (0.04,1.692) (0.05,1.652) (0.06,1.711) (0.07,1.567) (0.08,1.419) (0.09,1.275) (0.10,1.205) (0.11,1.102) (0.12,1.041) (0.13,0.901) (0.14,0.895) (0.15,0.791) (0.16,0.623) (0.17,0.468) (0.18,0.414) (0.19,0.326) (0.20,0.247) (0.21,0.230) (0.22,0.265) (0.23,0.224) (0.24,0.194) (0.25,0.168) (0.26,0.117) (0.27,0.123) (0.28,0.128) (0.29,0.170) (0.30,0.128) (0.31,0.090) (0.32,0.064) (0.33,0.014) (0.34,0.061) (0.35,0.050) (0.36,0.008) (0.37,0.031) (0.38,0.048) (0.39,0.028) (0.40,0.040) (0.41,0.026) (0.42,-0.011) (0.43,-0.016) (0.44,-0.023) (0.45,-0.015) (0.46,0.005) (0.47,-0.016) (0.48,0.005) (0.49,-0.027) (0.50,-0.038) (0.51,-0.042) (0.52,-0.071) (0.53,-0.041) (0.54,-0.034) (0.55,-0.019) (0.56,-0.007) (0.57,0.035) (0.58,0.053) (0.59,0.036) (0.60,0.040) (0.61,0.015) (0.62,0.041) (0.63,0.015) (0.64,-0.013) (0.65,-0.035) (0.66,-0.059) (0.67,-0.064) (0.68,-0.075) (0.69,-0.091) (0.70,-0.081) (0.71,-0.087) (0.72,-0.086) (0.73,-0.079) (0.74,-0.076) (0.75,-0.050) (0.76,-0.049) (0.77,-0.038) (0.78,-0.046) (0.79,-0.047) (0.80,-0.033) (0.81,-0.018) (0.82,-0.004) (0.83,-0.010) (0.84,-0.015) (0.85,-0.015) (0.86,-0.027) (0.87,-0.029) (0.88,-0.034) (0.89,-0.036) (0.90,-0.033) (0.91,-0.034) (0.92,-0.032) (0.93,-0.031) (0.94,-0.028) (0.95,-0.024) (0.96,-0.014) (0.97,-0.006) (0.98,0.005) (0.99,0.022) (1.00,0.000)};
\addlegendentry{8.3 million}
\end{axis}
\end{tikzpicture}
\end{minipage}
\caption{\textbf{In the Criteo experiment, effect estimates pay off in narrow campaigns.} (A) Average effect on visits by baseline group in the held-out data, with 95\% intervals, and the effect implied by a common shift fitted to treated users; 1.8\% of users have a baseline above 50\% (dotted). (B) Incremental visits per 1{,}000 users from ranking by the X-learner rather than by estimated BATE, at each treatment fraction (log scale), for experiments with 3{,}000 to 8.3 million treated users.}
\label{fig:criteo}
\end{figure}

The sizes above are those from which the gain, averaged over repeated experiments, stays positive at every larger size tested; they do not mark statistical significance. The held-out data add their own noise. With rankings trained on one experiment of a million treated users, the X-learner's gains among the top 1--10\% are well outside their standard errors and its Qini gain is three standard errors, while its differences among the top half and beyond are within one (\cref{app:criteo}).

% =====================================================================
\section{Discussion}\label{sec:discussion}
% =====================================================================

When ranking by estimated treatment effects improves on ranking by baseline predictions, it is tempting to conclude that the experiment uncovered differences in responsiveness that the baseline missed. The gain can arise even when people with the same baseline have exactly the same effect. In that case, all the experiment has done is show us how to put people with different baselines in a better order. Estimated BATE prioritization captures this first source of improvement, so what CATE prioritization adds beyond it is the second. Distinguishing the two helps explain both the mixed evidence on targeting and the very different experiment sizes needed to improve it. \Cref{tab:map} collects the main results as five insights, with what drives each and where it stops holding.

\begin{table}[t]
\centering
\caption{\textbf{Five insights: what drives each conclusion, and where it stops holding.} Each row gives a conclusion and the result that establishes it, the conditions behind it, and the conditions under which it fails or reverses.}
\label{tab:map}
\vspace{4pt}
\footnotesize
\renewcommand{\arraystretch}{1.1}
\begin{tabular}{@{}>{\raggedright\arraybackslash}p{5.6cm} >{\raggedright\arraybackslash}p{4.6cm} >{\raggedright\arraybackslash}p{5.0cm}@{}}
\toprule
Conclusion & What drives it & Where it stops holding \\
\midrule
\multicolumn{3}{@{}l}{\textit{1. Outcome predictions are competitive for a reason}} \\
The baseline can reveal all differences in effects (\cref{prop:sufficiency}) & A binary outcome crossing a threshold, with a common shift & Shifts that vary in ways the baseline cannot predict \\
The baseline is recovered better than effects (\cref{prop:amplify}) & Baselines vary more than effects, and bias aligned with effects & Strong treatments that more than double the odds of the outcome at low baselines \\
\addlinespace
\multicolumn{3}{@{}l}{\textit{2. Information is not ordering}} \\
The baseline can rank poorly despite complete potential; ranking by distance from the effect peak is optimal at every fraction (\cref{prop:onenumber}) & An effect peak inside the population & Everyone on one side of the peak, where baseline prioritization is already optimal \\
\addlinespace
\multicolumn{3}{@{}l}{\textit{3. Fixing the order can be cheap}} \\
Theory gets close without an experiment (\cref{cor:threshold}) & Symmetric noise and a modest shift & Large shifts; asymmetric noise \\
Errors in locating the peak cost little: the loss is quadratic in the error (\cref{prop:secondorder}) & One parameter, learned from everyone; effects are flat near the peak & A baseline that is not well estimated \\
\addlinespace
\multicolumn{3}{@{}l}{\textit{4. Forgone heterogeneity, not total heterogeneity, sets the experiment needed}} \\
The experiment needed grows with the ratio of captured to forgone heterogeneity (\cref{prop:unseen}, \cref{eq:nstar}) & Forgone heterogeneity against estimation error & Sizes are approximate and depend on the estimator \\
\addlinespace
\multicolumn{3}{@{}l}{\textit{5. Information beyond the baseline earns its value through the decisions it changes}} \\
CATE prioritization beats BATE prioritization at a capacity when forgone heterogeneity at the cutoff exceeds the error there (\cref{thm:margin}) & Departures from the BATE change decisions only near the cutoff & Large departures, which widen the band of people who can change sides \\
Narrow campaigns often need less (\cref{cor:narrow}) & Forgone heterogeneity grows faster than outcome noise with sensitivity, and the most sensitive are ranked first & Unpredictable shifts much larger at the bottom of the ranking \\
\bottomrule
\end{tabular}
\end{table}

Both sources of improvement appear in the literature. In the financial-aid experiment of \citet{athey2025}, effects peaked at intermediate baselines and varied little beyond what the baseline captured. A logit with one treatment coefficient outperformed a causal forest, consistent with our results in \cref{sec:enough}. In the retention campaigns of \citet{ascarza2018}, the highest-risk customers almost always churned and proved unresponsive, consistent with being far below the effect peak for staying, where a lower chance of staying means a smaller effect. But risk and responsiveness were also nearly unrelated, and the intervention backfired for some customers. Those findings suggest a second use for experimental data in that setting: learning which customers respond differently even when their risk is similar.

For practitioners, the first question is whether the baseline is being used well. In the threshold model with a common shift, if everyone lies on one side of the effect peak, ranking by the baseline in the matching direction is already optimal. If the population spans the peak, the ordering needs attention. With a modest common shift and symmetric noise, ranking people by how close their baseline is to 50\% gives a useful approximation, provided the probabilities are reasonably calibrated. When untreated data pin down the baseline, a small treated sample can locate the peak more precisely.

The second question is what the baseline misses. People with the same baseline may respond differently because treatment appeals to a motive the baseline does not capture. When departures from the BATE are small, learning those differences changes treatment decisions only near the cutoff, and it pays only if effects vary more there than the error in estimating them. The natural check is therefore to compare CATE prioritization with estimated BATE prioritization on held-out experimental data at the treatment fraction where the ranking will be used, rather than averaged across fractions \citep{yadlowsky2025rate}. A shrinkage estimator can also put partial weight on estimated departures from the BATE; how much weight to give them at each capacity remains an open question.

How much should an organization spend to improve its ranking? When experimentation and treatment draw on the same budget, a larger experiment leaves fewer people to treat, and the gain from a better ranking over the deployment horizon must cover the benefit forgone among them. Progressive sampling offers a way to stage that investment \citep{provost1999progressive}, with anytime-valid confidence sequences keeping repeated comparisons honest \citep{howard2021confidence}. When a ranking's loss shrinks like $1/n$, each doubling of the experiment buys half the improvement of the one before, at twice the cost. Yet a small pilot can favor the ranking that eventually loses: in \cref{fig:straddle}B, CATE prioritization trails estimated BATE prioritization until about 1{,}600 treated people, and in the advertising experiment, among the top 25\%, the X-learner trails until about 300{,}000 treated users. Even when a larger experiment is hard to justify, a smaller one can earn its cost by showing how to use the baseline and how large effects are, which untreated data alone cannot reveal.

We focus on binary outcomes and a single treatment because the threshold model gives us a tractable way to study both learning tasks in the same setting. Some conclusions depend on stronger assumptions (\cref{tab:map}, last column). Fixing the order can require little experimental data when the link is known, the shift is common, and the baseline is well estimated. The experiment-size formulas are Gaussian or local approximations, which we check in the running example. The advertising experiment illustrates how the gains can vary with capacity, although we cannot observe forgone heterogeneity directly to establish the mechanism.

But baseline predictions are not the only scores organizations already hold. They also rank by predictions of other outcomes, such as engagement or lifetime value, and by effects estimated in earlier campaigns. Potential, recovery, monotonicity, and bias alignment apply to any score, and so does the trade-off at the cutoff (\cref{thm:margin}), with the score's expected effect in place of the BATE. What is specific to the baseline is the threshold structure that tells us how effects vary with it, which makes the order cheap to fix. For another score, that relationship must be learned, though it is still a function of one variable. Learning it, as BATE prioritization does for the baseline, is a form of causal post-processing, which uses limited experimental data to adapt existing predictive scores to a treatment decision \citep{fernandez2025postprocessing}. 

We see this as a promising direction for future research. Organizations hold many predictions that were not built for treatment decisions, and putting them to better use with limited experimental data may often be more practical than learning effects from scratch. The two learning tasks studied here offer a starting point: they separate how well a score is used from what it misses, and they show why the value of learning more depends on the treatment capacity.

\section*{Code and Data}

\ifORsubmission
Code that reproduces every simulation, table, figure, and number in the paper, with a README, is provided in a zip file uploaded with the submission. The Criteo data are publicly available \citep{diemert2021benchmark} under a noncommercial license and must be downloaded separately; the README explains how.
\else
Code that reproduces every simulation, table, figure, and number in the paper, with a README, is available at \url{https://github.com/ferlocar/cate-baseline-prioritization}. The Criteo data are publicly available \citep{diemert2021benchmark} under a noncommercial license and must be downloaded separately; the README explains how.
\fi

% =====================================================================
\bibliographystyle{plainnat}
\bibliography{reference}

\ifORsubmission
  \processdelayedfloats
  \let\table\origtable \let\endtable\endorigtable
\fi
\clearpage
\appendix
\ifORsubmission
\setcounter{section}{0}
\renewcommand{\thesection}{EC.\arabic{section}}
% Keep e-companion PDF destinations distinct after resetting counters.
\def\theHfigure{EC.\arabic{figure}}
\def\theHtable{EC.\arabic{table}}
\def\theHproposition{EC.\arabic{proposition}}
\def\theHcorollary{EC.\arabic{corollary}}
\def\theHlemma{EC.\arabic{lemma}}
\def\theHassumption{EC.\arabic{assumption}}
\def\theHexample{EC.\arabic{example}}
\renewcommand{\theequation}{EC.\arabic{equation}}\setcounter{equation}{0}
\renewcommand{\thefigure}{EC.\arabic{figure}}\setcounter{figure}{0}
\renewcommand{\thetable}{EC.\arabic{table}}\setcounter{table}{0}
\renewcommand{\theproposition}{EC.\arabic{proposition}}\setcounter{proposition}{0}
\renewcommand{\thecorollary}{EC.\arabic{corollary}}\setcounter{corollary}{0}
\renewcommand{\thelemma}{EC.\arabic{lemma}}\setcounter{lemma}{0}
\renewcommand{\theassumption}{EC.\arabic{assumption}}\setcounter{assumption}{0}
\renewcommand{\theexample}{EC.\arabic{example}}\setcounter{example}{0}
\begin{center}
{\Large\bfseries E-Companion\par}
\end{center}
\else
\counterwithin{lemma}{section}
\fi
\setlength{\emergencystretch}{2em}

% =====================================================================
\section{Proofs}\label{app:proofs}
% =====================================================================

Throughout the proofs, $0<\Var(\beta)<\infty$ and $m_S=\E[\beta\mid S]$. Expected effects may be constant over ranges of a score. Scores may have atoms, as in the discrete examples or when ranking by $m_\theta$; ties at the cutoff are then broken at random, independently of effects, as in \cref{sec:setup}.

\subsection{Tools}

For an integrable random variable $A$ with CDF $F_A$, let $Q_A(p)=\inf\{t\in\R:F_A(t)\ge p\}$ be its quantile function, $L_A(t)=\E[(A-t)_+]$ its stop-loss function, and $H_A(u)=\int_{1-u}^1Q_A(p)\,dp$ the total of its largest fraction $u$ of values, per member of the population. We write $A\cx B$ if $\E[\phi(A)]\ge\E[\phi(B)]$ for every convex $\phi$.

\begin{lemma}[Convex order]\label{lem:stoploss}
For integrable $A,B$ with equal means, $A\cx B$ if and only if $L_A(t)\ge L_B(t)$ for all $t$, if and only if $H_A(u)\ge H_B(u)$ for all $u\in[0,1]$.
\end{lemma}

These equivalences are standard \citep[Section~3.A]{shaked2007stochastic}.

Since $\mathbf 1\{F_S(S)\ge1-u\}$ is a function of $S$, conditioning on $S$ gives $V(u;S)=\E\big[\E[\beta\mid S]\,\mathbf 1\{F_S(S)\ge1-u\}\big]=\E[m_S\,\mathbf 1\{F_S(S)\ge1-u\}]$. Let $q=Q_{m_S}(1-u)$. Since $Q_{m_S}(p)\le q$ for $p<1-u$ and $Q_{m_S}(p)\ge q$ for $p>1-u$, $H_{m_S}(u)=qu+\E[(m_S-q)_+]$. Hence, for any random variable $W\in[0,1]$ with $\E[W]=u$,
\[
\E[m_S W]=qu+\E[(m_S-q)W]\le qu+\E[(m_S-q)_+]=H_{m_S}(u).
\]
Taking $W=\mathbf 1\{F_S(S)\ge1-u\}$ gives
\begin{equation}\label{eq:topu}
V(u;S)\le H_{m_S}(u).
\end{equation}
Equality holds for all $u$ if $S$ satisfies signal monotonicity: with $U=F_S(S)$ uniform, $m_S(Q_S(U))$ is a nondecreasing function of $U$ with the distribution of $m_S$, so it equals $Q_{m_S}(U)$ almost surely. With atoms, randomized ties give $V(u;S)=\E[m_S\,w_u(S)]$ for weights $w_u\in[0,1]$, nondecreasing in $S$, with $\E[w_u(S)]=u$; taking $W=w_u(S)$ above gives \eqref{eq:topu}, with equality under signal monotonicity because $m_S$ is constant on each atom.

\begin{lemma}[Variance and convex order]\label{lem:variance}
Let $A,B$ be square-integrable random variables with CDFs $F_A$, $F_B$, and with the same mean.
\begin{enumerate}[label=(\alph*)]
  \item If $A\cx B$, then $\Var(A)\ge\Var(B)$.
  \item If $F_A(t)-F_B(t)$ changes sign at most once, then $\Var(A)\ge\Var(B)$ implies $A\cx B$, and $\Var(A)=\Var(B)$ implies that $A$ and $B$ have the same distribution.
\end{enumerate}
\end{lemma}

\begin{proof}
(a) Take $\phi(a)=a^2$, using $\E[A]=\E[B]$.

(b) Let $D(t)=F_A(t)-F_B(t)$, so $\int_{-\infty}^{\infty} D(t)\,dt=\E[B]-\E[A]=0$, since they have the same means, and from $L_A(z)=\int_z^\infty(1-F_A(t))\,dt$, $L_A(z)-L_B(z)=-\int_z^\infty D(t)\,dt=\int_{-\infty}^zD(t)\,dt$. Since $D$ changes sign at most once, there exists $c\in\R$ such that either $D(t)\ge 0$ for all $t<c$ and $D(t)\le 0$ for all $t>c$, or the reverse. In the first case, $\int_{-\infty}^z D(t)\,dt$ rises from $0$ up to $c$ and falls back to $0$, so $L_A(z)\ge L_B(z)$ for all $z$ and $A\cx B$ by \cref{lem:stoploss}; in the second case, $B\cx A$. For the variances, equal means give $\Var(A)-\Var(B)=\E[(A-c)^2]-\E[(B-c)^2]$. Writing $(A-c)^2=\int_c^\infty 2(t-c)\mathbf 1\{A>t\}\,dt+\int_{-\infty}^c 2(c-t)\mathbf 1\{A<t\}\,dt$, taking expectations, and subtracting the same expression for $B$ yields $\Var(A)-\Var(B)=-2\int_{-\infty}^{\infty}(t-c)D(t)\,dt$. Thus the first case has $\Var(A)\ge\Var(B)$ and the second $\Var(A)\le\Var(B)$, each strictly unless $D=0$ almost everywhere \citep[see also][]{ohlin1969class}.
\end{proof}

\begin{lemma}[Information loss]\label{lem:garbling}
If $\E[\beta\mid S_1,S_2]=\E[\beta\mid S_1]$, then $m_{S_1}\cx m_{S_2}$. In particular, $\beta\cx m_S$ for every score $S$.
\end{lemma}

\begin{proof}
By iterated expectations, $m_{S_2}=\E[m_{S_1}\mid S_2]$, so conditional Jensen's inequality gives $\E[\phi(m_{S_2})]\le\E[\phi(m_{S_1})]$ for convex $\phi$. For the last claim, take $S_1=X$, so $m_{S_1}=\beta$.
\end{proof}

\subsection{Proofs for \texorpdfstring{\cref{sec:setup}}{Section 2}}

\begin{proof}[Proof of \cref{prop:decomp}]
For any square-integrable $A$ and score $S$, iterated expectations give $\Cov(A,\E[A\mid S])=\Var(\E[A\mid S])$, so $\Corr(A,\E[A\mid S])^2=\Var(\E[A\mid S])/\Var(A)$. The assumption gives $m_{\hat q}=\E[m_q\mid\hat q]$, so taking $A=m_q$ and $S=\hat q$ gives $\rho^2_q=\Var(m_{\hat q})/\Var(m_q)$. Multiplying by $R^2_{\beta\mid q}=\Var(m_q)/\Var(\beta)$ proves the decomposition. (If $m_q$ or $m_{\hat q}$ is constant, $\rho_q=0$; both sides are then zero.) For an effect estimate, take $A=\beta$ and $S=\hat\beta$.
\end{proof}

\begin{proof}[Proof of \cref{prop:bar}]
Signal monotonicity and \eqref{eq:topu} give $V(u;S_1)=H_{m_{S_1}}(u)$ for every $u$.

(a) \Cref{lem:garbling} gives $m_{S_1}\cx m_{S_2}$, so $H_{m_{S_1}}(u)\ge H_{m_{S_2}}(u)\ge V(u;S_2)$ by \cref{lem:stoploss} and \eqref{eq:topu}. \Cref{lem:variance}(a) gives $\Var(m_{S_1})\ge\Var(m_{S_2})$.

(b) Now also $V(u;S_2)=H_{m_{S_2}}(u)$, so uniform dominance is equivalent to $m_{S_1}\cx m_{S_2}$ (\cref{lem:stoploss}), which under single crossing is equivalent to $\Var(m_{S_1})\ge\Var(m_{S_2})$ (\cref{lem:variance}).
\end{proof}

\subsection{Proofs for \texorpdfstring{\cref{sec:threshold}}{Section 3}}

\begin{lemma}[Shape of the effect]\label{lem:shape}
With a common shift $\delta>0$, $\beta(\mu)=F(\mu+\delta)-F(\mu)$ is symmetric about $-\delta/2$, strictly increasing on $(-\infty,-\delta/2]$, and strictly decreasing on $[-\delta/2,\infty)$.
\end{lemma}

\begin{proof}
Symmetry of the noise gives $\beta(-\delta-\mu)=F(-\mu)-F(-\delta-\mu)=\beta(\mu)$. The derivative is $\beta'(\mu)=f(\mu+\delta)-f(\mu)$, which is positive when $|\mu+\delta|<|\mu|$, that is, when $\mu<-\delta/2$, and negative when $\mu>-\delta/2$.
\end{proof}

So under a common shift, $\beta=\psi(|\mu+\delta/2|)$ for a strictly decreasing $\psi$.

\begin{proof}[Proof of \cref{prop:sufficiency}]
Since $f$ is strictly decreasing in $|z|$, it is positive everywhere, so $F$ is a strictly increasing bijection from $\R$ to $(0,1)$, so $\mu=F^{-1}(\theta)$ and $\beta=F(F^{-1}(\theta)+\delta)-\theta=h(\theta)$. The shape of $h$ follows from \cref{lem:shape}, with the peak at $\theta^*=F(-\delta/2)$. Since $\beta$ is a function of $\theta$, $m_\theta=\beta$ and $R^2_{\beta\mid\theta}=1$.
\end{proof}

\begin{lemma}[One side of the peak]\label{lem:oneside}
Under a common shift, if everyone is on one side of the peak, ranking by the baseline in the matching direction uniformly dominates every score.
\end{lemma}

\begin{proof}
By \cref{lem:shape}, the ranking orders people by a strictly increasing function of $\beta$, so its expected effect is $\beta$ itself, it satisfies signal monotonicity, and its curve is $H_\beta(u)$ by \eqref{eq:topu}. Any other score $S$ has $V(u;S)\le H_{m_S}(u)\le H_\beta(u)$ by \eqref{eq:topu} and \cref{lem:garbling,lem:stoploss}.
\end{proof}

\begin{proof}[Proof of \cref{prop:amplify}]
(a) Since $e_q$ and $e_\beta$ have the same conditional distribution given $(\beta,q)$, and the bias $\alpha=q-\beta$ is a function of $(\beta,q)$, the pair $(\hat q,\beta)$ has the same distribution as $(\hat\beta+\alpha,\beta)$. A score's prioritization curve depends only on its joint distribution with $\beta$, so $V(u;\hat q)=V(u;\hat\beta+\alpha)$, and we may take $\hat q=\hat\beta+\alpha$. Fix $0<u<1$. Call \emph{added} the people selected by $\hat q$ but not by $\hat\beta$, and \emph{dropped} those selected by $\hat\beta$ but not by $\hat q$. Both groups have the same probability $p_u$: each ranking selects a fraction $u$ of the population, so each group has probability $u$ minus the probability of being selected by both. Thus,
\begin{equation}
\begin{aligned}
&V(u;\hat q)-V(u;\hat\beta)\\
&\quad=\E\big[\beta\mid F_{\hat q}(\hat q)\ge 1-u,\,F_{\hat\beta}(\hat\beta)<1-u\big]\,\mathbb{P}\big(F_{\hat q}(\hat q)\ge 1-u,\,F_{\hat\beta}(\hat\beta)<1-u\big)\\
&\qquad-\E\big[\beta\mid F_{\hat q}(\hat q)<1-u,\,F_{\hat\beta}(\hat\beta)\ge 1-u\big]\,\mathbb{P}\big(F_{\hat q}(\hat q)<1-u,\,F_{\hat\beta}(\hat\beta)\ge 1-u\big)\\
&\quad=p_u\big\{\E[\beta\mid\text{added}]-\E[\beta\mid\text{dropped}]\big\}.
\end{aligned}
\end{equation}
For an added $i$ and a dropped $j$, $\hat\beta_i<\hat\beta_j$ and $\hat q_i>\hat q_j$, so $\alpha_i>\alpha_j$. If $\beta_i<\beta_j$, alignment would give $\alpha_i\le\alpha_j$. So every added person has at least the effect of every dropped person, and the difference is nonnegative. 

(b) By part (a), it suffices to check alignment. On the falling side, $\beta$ decreases in $\mu$ (\cref{lem:shape}), and so does the bias of $q=-\theta$, $\alpha=-\theta-\beta=-F(\mu+\delta)$; larger effects therefore come with larger bias. On the rising side, $\beta$ increases in $\mu$, and the bias of $q=\theta$ is $\alpha=2F(\mu)-F(\mu+\delta)$, with derivative $2f(\mu)-f(\mu+\delta)$. If $f(z+\delta)\le2f(z)$ on the interval spanned by the support, $\alpha$ is nondecreasing in $\mu$ and hence in $\beta$. (Where the support has gaps, the condition must also hold across them.)
\end{proof}

\paragraph{More on amplification.}\phantomsection\label{app:amplify}
Does a biased score help simply because a larger target stands out more from the same noise? No: a departure unrelated to effects hurts.

\begin{proposition}[Unrelated departures hurt]\label{prop:unrelated}
Let $S=\hat\beta+\nu$, where $\nu$ is independent of $(\beta,\hat\beta)$. If $\hat\beta$ satisfies signal monotonicity, then $\hat\beta$ uniformly dominates $S$.
\end{proposition}

\begin{proof}
Independence gives $\E[\beta\mid\hat\beta,S]=\E[\beta\mid\hat\beta]$, so \cref{prop:bar}(a) applies.
\end{proof}

What helps is a larger score in the right places, which is what alignment requires. On the rising side, \cref{prop:amplify}(b) asks that $f(z+\delta)\le2f(z)$: treatment may at most double the outcome's sensitivity to the index. For rare outcomes, this ratio approaches the relative risk $\mathbb P(Y(1)=1\mid X)/\mathbb P(Y(0)=1\mid X)$. If $f$ is log-concave, the ratio $f(z+\delta)/f(z)$ decreases in $z$ and equals one at the peak, so the condition holds on the interval spanned by the support if and only if it holds at the lowest index, or in the limit if the support is unbounded below. The statements below refer to that interval.

\begin{corollary}[Logistic and Gaussian noise]\label{cor:links}
On the rising side:
\begin{enumerate}[label=(\alph*)]
  \item With logistic noise, the bias is aligned everywhere if $\delta\le\log2$, that is, if treatment at most doubles the odds of the outcome. Otherwise, it is aligned where $\theta\ge(\sqrt{e^{\delta}/2}-1)/(e^{\delta}-1)$.
  \item With Gaussian noise, for every $\delta>0$ the bias is aligned exactly where $\theta\ge\Phi(-\delta/2-\log2/\delta)$.
\end{enumerate}
\end{corollary}

\begin{proof}
Both densities are log-concave. (a) With $p=F(z)$ and $a=e^\delta$, $F(z+\delta)=ap/\{1+(a-1)p\}$ and $f=F(1-F)$, so $f(z+\delta)/f(z)=a/\{1+(a-1)p\}^2$, which tends to $a$ as $p\to0$ and is at most 2 if and only if $p\ge(\sqrt{a/2}-1)/(a-1)$. (b) $\phi(z+\delta)/\phi(z)=\exp(-\delta z-\delta^2/2)$, which is at most 2 if and only if $z\ge-\delta/2-\log2/\delta$.
\end{proof}

With logistic noise, a common shift multiplies the odds by a constant, so at low baselines the effect is roughly a constant fraction of the baseline. With Gaussian noise, the relative lift from a fixed shift grows without bound as the baseline falls, so for rare enough outcomes the effect changes faster than the baseline does. With Gaussian noise and $\delta=0.3$, for example, alignment fails below a baseline of about 0.7\%.

\subsection{Proofs for \texorpdfstring{\cref{sec:enough}}{Section 4}}

\begin{proof}[Proof of \cref{prop:onenumber}]
Since $\beta=\psi(|\mu+\delta/2|)$ with $\psi$ strictly decreasing, ranking by $-|\mu+\delta/2|$ is ranking by a strictly increasing function of $\beta$, and the argument of \cref{lem:oneside} applies.
\end{proof}

\begin{proof}[Proof of \cref{prop:secondorder}]
Denote $\mu^*=-\delta/2$, $g$ the density of $\mu$, and $G$ the corresponding CDF. Fix $0<u<1$ (at $u=1$ both rankings treat everyone). Ranking by distance from a center $z$ treats the interval of probability $u$ closest to $z$: writing $\ell<\bar\ell$ for the endpoints of the support, it treats $[a(z),b(z)]$ with $a(z)=\max\{\ell,z-r(z)\}$ and $b(z)=\min\{\bar\ell,z+r(z)\}$, where $[z-r(z),z+r(z)]$ has probability $u$. Its value is $W(z)=\int_{a(z)}^{b(z)}\beta g$, and the loss is $W(\mu^*)-W(\mu^*+d)\ge0$ by \cref{prop:onenumber}.

\emph{Both edges interior.} Then $a+b=2z$ and $G(b)-G(a)=u$. Differentiating,
\begin{equation}
a'=\frac{2g(b)}{g(a)+g(b)},\qquad b'=\frac{2g(a)}{g(a)+g(b)},
\end{equation}
so
\begin{equation}
W'(z)=\beta(b)g(b)b'-\beta(a)g(a)a'=J(z)\big\{\beta(b)-\beta(a)\big\},\qquad J(z)=\frac{2g(a)g(b)}{g(a)+g(b)}\le\bar g.
\end{equation}
Here $J(z)\le\bar g$ because it is the harmonic mean of $g(a)$ and $g(b)$. By symmetry of $\beta$ about $\mu^*$, $\beta(a)=\beta(2\mu^*-a)$, and $b-(2\mu^*-a)=2(z-\mu^*)$. Since $|\beta'|\le L_\delta$, $|\beta(b)-\beta(a)|\le2L_\delta|z-\mu^*|$, so $|W'(z)|\le2\bar gL_\delta|z-\mu^*|$.

\emph{One edge clipped.} If $a=\ell$, then $b=G^{-1}(u)$ does not depend on $z$, and similarly if $b=\bar\ell$; so $W$ is constant. Both edges cannot be clipped when $u<1$.

\emph{The loss.} The edges are continuous in $z$, with derivatives in $[0,2]$ where both are interior, so $W$ is Lipschitz on compact intervals ($\beta g$ is bounded) and the bound $|W'(z)|\le2\bar gL_\delta|z-\mu^*|$ holds almost everywhere. Integrating from $\mu^*$ to $\mu^*+d$ gives $W(\mu^*)-W(\mu^*+d)\le\bar gL_\delta d^2$.
\end{proof}

\begin{proof}[Proof of \cref{cor:threshold}]
By symmetry, $|F(\mu)-1/2|=F(|\mu|)-1/2$, so threshold prioritization ranks by $-|\mu|$, the distance from $\mu^*+d$ with $d=\delta/2$. \Cref{prop:secondorder} gives the bound. For the claim in the main text: if every baseline is at most $\theta^*$, then $\mu<0$ for everyone and $-|\mu|=\mu$, the matching direction on the rising side; if every baseline is at least $1/2$, then $-|\mu|=-\mu$, the matching direction on the falling side.
\end{proof}

The bound is small relative to the value at stake. If $M=\sup_z|f'(z)|<\infty$, then $L_\delta\le M\delta$, so the loss is $O(\delta^3)$, while $\beta=\delta f(\mu)+O(\delta^2)$ makes the best value at a fixed fraction of order $\delta$.

\subsection{Proofs for \texorpdfstring{\cref{sec:size}}{Section 5}}\label{app:proofs5}

Equation \eqref{eq:potential} is the law of total variance, using that $\theta=F(\mu)$ and $\mu$ carry the same information.

\begin{proof}[Proof of \cref{prop:unseen}]
(a) If $\eta=0$, then $\beta=F(\mu+\E[\delta\mid\mu])-F(\mu)$ is a function of $\mu$, so $\mathcal U=\E[\Var(\beta\mid\mu)]=0$.

(b) Let $a=\mu+\E[\delta\mid\mu]$ and $M=\sup_z|f'(z)|$. Since $\theta$ and $\mu$ carry the same information, $\mathcal U=\E[\Var(\beta\mid\mu)]$. Taylor's theorem around $a$ gives
\begin{equation}
\beta=F(a+tZ)-F(\mu)=\big\{F(a)-F(\mu)\big\}+tf(a)Z+R,\qquad|R|\le\tfrac M2t^2Z^2.
\end{equation}
Given $\mu$, $F(a)-F(\mu)$ and $tf(a)$ are fixed numbers, so
\begin{equation}
\Var(\beta\mid\mu)=t^2f(a)^2\Var(Z\mid\mu)+2tf(a)\Cov(Z,R\mid\mu)+\Var(R\mid\mu).
\end{equation}
Since $\E[Z\mid\mu]=0$, $\Cov(Z,R\mid\mu)=\E[ZR\mid\mu]$, so with $f\le\bar f=f(0)$ the middle term is at most $\bar fMt^3\E[|Z|^3\mid\mu]$ in absolute value. The last is at most $\E[R^2\mid\mu]\le\tfrac{M^2}{4}t^4\E[Z^4\mid\mu]$. Taking expectations, and using $t^2\Var(Z\mid\mu)=\Var(\eta\mid\mu)$ and $\E|Z|^3\le\E[Z^4]^{3/4}<\infty$, gives $\mathcal U=\E[f(a)^2\Var(\eta\mid\mu)]+O(t^3)$.
\end{proof}

\Cref{thm:margin} is stated in the setting of \cref{sec:narrow}, where the BATE is monotone in the baseline, so baseline prioritization ranks as BATE prioritization does and $V(u;\theta)=V(u;m_\theta)$; the proposition itself holds for BATE prioritization without this assumption. Effects are $\beta=m_\theta+t\zeta$ with $t>0$ and $\E[\zeta\mid\theta]=0$, and the estimate is $\hat\beta=m_\theta+t\xi$ with $\E[\hat\beta\mid\theta]=m_\theta$, that is, $\E[\xi\mid\theta]=0$. The prioritization curve of $\hat\beta$ and its cutoff are computed over people and estimation errors together. This describes estimators whose errors are independent across many cells, such as the cell estimator, for which the ranking from a single experiment is close to this average. The proposition also uses four regularity conditions, which rule out heavy-tailed departures and discontinuities in the population at the cutoff:
\begin{enumerate}[label=(R\arabic*)]
  \item $\zeta$ is bounded;
  \item $|\xi|\le\Xi$ for a random variable $\Xi$ independent of $\theta$ with $\E[\Xi^2]<\infty$;
  \item $m_\theta$ has a density $g_{m_\theta}$ that is continuous and positive near $c_u$;
  \item the joint distribution of $(\zeta,\xi)$ conditional on $m_\theta$ varies continuously with $m_\theta$ near $c_u$.
\end{enumerate}

\begin{proof}[Proof of \cref{thm:margin}]
Write $g$ for the density $g_{m_\theta}$ of $m_\theta$, $c=c_u$, and $\hat c$ for the upper $u$-quantile of $\hat\beta=m_\theta+t\xi$, with $h=\hat c-c$. Since $m_\theta$ is a function of $\theta$, $\E[\zeta\mid m_\theta]=\E[\E[\zeta\mid\theta]\mid m_\theta]=0$, and similarly, $\E[\xi\mid m_\theta]=0$.

Fix $\varepsilon>0$ so that (R3)--(R4) hold and $g$ is bounded on the window $[c-\varepsilon,c+\varepsilon]$. By (R2), $\mathbb P(|t\xi|>\eta)\le\mathbb P(\Xi>\eta/t)\to0$ as $t\to0^+$ for any $\eta>0$, and since $g(c)>0$ by (R3), $\hat c\to c$, so $|\hat c-c|<\varepsilon/2$ for small $t$. Any person outside this window has $|m_\theta-c|>\varepsilon$, and is therefore more than $\varepsilon/2$ away from $\hat c$. For such a person's score $\hat\beta=m_\theta+t\xi$ to switch sides of the cutoff, the perturbation must bridge this gap: $|t\xi|>\varepsilon/2$, which by (R2) implies $\Xi\ge|\xi|>\varepsilon/(2t)$.

Since $\mathbf 1\{\Xi>a\}\le(\Xi^2/a^2)\mathbf 1\{\Xi>a\}$ for $a=\varepsilon/(2t)$, taking expectations gives
\begin{equation}
\mathbb{P}(\Xi>\varepsilon/(2t))
\le
\frac{4t^2}{\varepsilon^2}\,\E\big[\Xi^2\mathbf 1\{\Xi>\varepsilon/(2t)\}\big]=o(t^2),
\end{equation}
where the first inequality follows by Markov's inequality, and the second one using that $\E[\Xi^2]<\infty$. Since $m_\theta\in[-1,1]$ and $\zeta$ is bounded by (R1), people outside $[c-\varepsilon,c+\varepsilon]$ contribute $o(t^2)$ to all quantities below, so we restrict every integral to $|m_\theta-c|\le\varepsilon$ up to an error of $o(t^2)$.

\emph{Step 1: the cutoff barely moves.} Using the result above, and for $x$ near the cutoff $c$, by conditioning on $m_\theta=b$ and writing $b=x+ty$ for some $y$:
\begin{equation}
\begin{aligned}
\mathbb P(\hat\beta\ge x)-\mathbb P(m_\theta\ge x)
% &=\E\big[\mathbf 1\{m_\theta+t\xi\ge x\}-\mathbf 1\{m_\theta\ge x\}\big]\\
&=\int_{c-\varepsilon}^{c+\varepsilon}g(b)\big\{\mathbb P(b+t\xi\ge x\mid m_\theta=b)-\mathbf 1\{b\ge x\}\big\}\,db+o(t^2)\\
&=t\int_{(c-\varepsilon-x)/t}^{(c+\varepsilon-x)/t} g(x+ty)\big\{\mathbb P(\xi\ge-y\mid m_\theta=x+ty)-\mathbf 1\{y\ge0\}\big\}\,dy+o(t^2). \label{eq:cutoff-barely-moves}
\end{aligned}
\end{equation} 

Since $\mathbf 1\{\xi\ge-y\}$ and $\mathbf 1\{y\ge0\}$ can differ only if $|\xi|\ge|y|$, the integrand is bounded by a constant times $\mathbb P(\Xi\ge|y|)$, which is integrable. By dominated convergence, (R3)--(R4), and Fubini's theorem, as $t\to0^+$ along any sequence $x=x_t\to c$, since $\varepsilon$ is fixed, the limits of integration tend to $\pm\infty$ and the integral in \Cref{eq:cutoff-barely-moves} tends to
\begin{equation}
\begin{aligned}
g(c)\int_{-\infty}^{\infty}\big\{\mathbb P(\xi\ge-y\mid m_\theta=c)-\mathbf 1\{y\ge0\}\big\}\,dy
&=g(c)\,\E\!\left[\int_{-\infty}^{\infty}\big(\mathbf 1\{y\ge-\xi\}-\mathbf 1\{y\ge0\}\big)\,dy\,\Big|\,m_\theta=c\right]\\
&=g(c)\,\E[\xi\mid m_\theta=c]=0,
\end{aligned}
\end{equation}
because $\mathbf 1\{y\ge-\xi\}-\mathbf 1\{y\ge0\}$ equals $+1$ on $[-\xi,0)$ when $\xi>0$ and $-1$ on $[0,-\xi)$ when $\xi<0$, giving the step-function identity
\begin{equation}
\int_{-\infty}^{\infty}\big(\mathbf 1\{y\ge-\xi\}-\mathbf 1\{y\ge0\}\big)\,dy=\int_{-\xi}^{0}1\,dy=\xi.
\end{equation}
Hence $\mathbb P(\hat\beta\ge x_t)-\mathbb P(m_\theta\ge x_t)=o(t)$. Setting $x_t=\hat c$, since $\hat c$ is close to $c$, gives $\mathbb P(m_\theta\ge\hat c)=u+o(t)$, and subtracting $\mathbb P(m_\theta\ge c)=u$ yields
\begin{equation}
o(t)=\mathbb P(m_\theta\ge\hat c)-\mathbb P(m_\theta\ge c)=-\int_c^{\hat c}g(b)\,db=-\big(g(c)+o(1)\big)h,
\end{equation}
and $g(c)>0$ by (R3) gives $h=o(t)$. 

\emph{Step 2: decomposition.} By definition, $V(u;\hat\beta)=\E[\beta\,\mathbf 1\{\hat\beta\ge\hat c\}]$ and $V(u;m_\theta)=\E[\beta\,\mathbf 1\{m_\theta\ge c\}]$. Because both rankings treat a fraction $u$, $\chi=\E[c(\mathbf 1\{\hat\beta\ge\hat c\}-\mathbf 1\{m_\theta\ge c\})]=c(u-u)=0$. Substituting $\beta=m_\theta+t\zeta$ in the second line below, and subtracting the zero term $\chi$ in the third line, gives
\begin{equation}
\begin{aligned}
V(u;\hat\beta)-V(u;m_\theta)
&=\E\big[\beta\big(\mathbf 1\{\hat\beta\ge\hat c\}-\mathbf 1\{m_\theta\ge c\}\big)\big]\\
&= \E\big[m_\theta\big(\mathbf 1\{\hat\beta\ge\hat c\}-\mathbf 1\{m_\theta\ge c\}\big)\big] + t\,\E\big[\zeta\big(\mathbf 1\{\hat\beta\ge\hat c\}-\mathbf 1\{m_\theta\ge c\}\big)\big]\\
&=\underbrace{\E\big[(m_\theta-c)\big(\mathbf 1\{\hat\beta\ge\hat c\}-\mathbf 1\{m_\theta\ge c\}\big)\big]}_{A}+\underbrace{t\,\E\big[\zeta\big(\mathbf 1\{\hat\beta\ge\hat c\}-\mathbf 1\{m_\theta\ge c\}\big)\big]}_{D}. %%% we are dealing with enough c terms to add another one, changed it to D
\end{aligned}
\end{equation}

\emph{Step 3: the two terms.} For $A$, condition on $m_\theta=b$ and write $b=c+ty$. Since $h/t\to0$, we have $|h/t|\le1$ for small $t$, so as in Step~1 the integrand is bounded by a constant times $|y|\,\mathbb P(\Xi+1\ge|y|)$, which is integrable because $\E[\Xi^2]<\infty$. By dominated convergence, (R3)--(R4), and Fubini's theorem,
\begin{equation}
\begin{aligned}
A&=\int_{c-\varepsilon}^{c+\varepsilon}g(b)\,(b-c)\,\big\{\mathbb P(b+t\xi\ge\hat c\mid m_\theta=b)-\mathbf 1\{b\ge c\}\big\}\,db+o(t^2)\\
&=t^2\int_{-\varepsilon/t}^{\varepsilon/t} g(c+ty)\,y\,\big\{\mathbb P(\xi\ge h/t-y\mid m_\theta=c+ty)-\mathbf 1\{y\ge0\}\big\}\,dy+o(t^2)\\
&=t^2g(c)\,\E\!\left[\int_{-\infty}^{\infty}y\big(\mathbf 1\{y\ge-\xi\}-\mathbf 1\{y\ge0\}\big)\,dy\,\Big|\,m_\theta=c\right]+o(t^2)\\
&=-\frac{t^2g(c)}{2}\,\E[\xi^2\mid m_\theta=c]+o(t^2),
\end{aligned}
\end{equation}
where the inner integral equals $\int_{-\xi}^{0}y\,dy=-\xi^2/2$ by the same step-function argument as in Step~1.

Similarly, for the term $D$,  %%% we are dealing with enough c terms to add another one
since $\zeta$ is bounded by (R1), the same substitution $b=c+ty$ and Step~1 identity yield
\begin{equation}
\begin{aligned}
D&=t^2\int_{-\varepsilon/t}^{\varepsilon/t} g(c+ty)\,\E\big[\zeta\big(\mathbf 1\{\xi\ge h/t-y\}-\mathbf 1\{y\ge0\}\big)\mid m_\theta=c+ty\big]\,dy+o(t^2)\\
&=t^2g(c)\,\E\!\left[\zeta\int_{-\infty}^{\infty}\big(\mathbf 1\{y\ge-\xi\}-\mathbf 1\{y\ge0\}\big)\,dy\,\Big|\,m_\theta=c\right]+o(t^2)\\
&=t^2g(c)\,\E[\zeta\xi\mid m_\theta=c]+o(t^2).
\end{aligned}
\end{equation}
Adding $A$ and $D$ and using $2\zeta\xi-\xi^2=\zeta^2-(\xi-\zeta)^2$,
\begin{equation}
V(u;\hat\beta)-V(u;m_\theta)=\frac{g(c)}{2}\Big\{t^2\E[\zeta^2\mid m_\theta=c]-t^2\E[(\xi-\zeta)^2\mid m_\theta=c]\Big\}+o(t^2).
\end{equation}
Since $\E[\zeta\mid m_\theta]=0$, the first term in braces is $\Var(\beta\mid m_\theta=c)$, and since $\hat\beta-\beta=t(\xi-\zeta)$, the second is $\E[(\hat\beta-\beta)^2\mid m_\theta=c]$.
\end{proof}

\begin{proof}[Proof of \cref{cor:narrow}]
Write $\delta=\bar\delta+\eta$ with $\eta=tZ$ and $\E[\eta\mid\mu]=0$. Expanding $F(\mu+\bar\delta+\eta)$ around $\mu+\bar\delta$ gives
\begin{equation}
\beta=F(\mu+\bar\delta)-F(\mu)+f(\mu+\bar\delta)\eta+O(t^2).
\end{equation} 
Since $\mathbb{E}[\eta\mid\mu]=0$,
\begin{equation}
m_\theta=\mathbb{E}[\beta\mid\mu]=F(\mu+\bar\delta)-F(\mu)+O(t^2).
\end{equation}
This is strictly monotone in $\mu$ on each side of the peak for small $t$, so conditioning on $m_\theta=c_u$ is conditioning on the index at the margin, $\mu_u$, which moves in one direction as $u$ grows. By \cref{prop:unseen}(b), $\Var(\beta\mid\mu)=vf(\mu+\bar\delta)^2$ to leading order, and with logistic noise, $F(1-F)=f$ gives $w=\theta_1(1-\theta_1)=f(\mu+\bar\delta+\eta)=f(\mu+\bar\delta)+f'(\mu+\bar\delta)\eta+O(t^2)$, so $\E[w\mid\mu]=f(\mu+\bar\delta)$ to leading order. So $n_{\mathrm{margin}}^*(u)$ in \eqref{eq:margin} is proportional to
\begin{equation}
r(\mu)=\frac{f(\mu+\bar\delta)}{v\,f(\mu+\bar\delta)^2}=\frac{1}{v\,f(\mu+\bar\delta)}
\end{equation}
at $\mu=\mu_u$. More generally, with $\Var(\eta\mid\mu)$ in place of $v$, $n_{\mathrm{margin}}^*(u)$ is nondecreasing exactly when $\Var(\eta\mid\mu)f(\mu+\bar\delta)$ is nonincreasing down the ranking. The logistic density $f$ is increasing below zero and decreasing above it.

On the falling side, $\mu>-\bar\delta/2$ gives $\mu+\bar\delta\ge \bar\delta/2>0$, so $1/f(\mu+\bar\delta)$ increases with $\mu$, and BATE prioritization, ranking from the lowest baseline up, moves $\mu_u$ up as $u$ grows.

On the rising side, BATE prioritization ranks from the highest baseline down, which moves $\mu_u$ down as $u$ grows. The condition $F(\mu+\bar\delta)\le1/2$ gives $\mu+\bar\delta\le0$, so $1/f(\mu+\bar\delta)$ is nonincreasing in $\mu$.

The statement concerns this leading-order expression. To apply \cref{thm:margin} itself as unpredictable shifts shrink, take $Z$ bounded; with the support away from the peak, $m_\theta$ then has a positive, continuous density near each interior cutoff, and the argument carries over.
\end{proof}

% =====================================================================
\section{The Gaussian Case}\label{app:gaussian}
% =====================================================================

This appendix develops the Gaussian approximation summarized in \cref{sec:thebar}. It is a case of \cref{prop:bar}(b) in which every quantity is Gaussian.

\begin{assumption}[Gaussian case]\label{ass:gaussian}
Effects $\beta$ and the baseline $\theta$ are jointly Gaussian with positive variances $\Var(\beta)$ and $\Var(\theta)$ and squared correlation $P$, so $P$ is the baseline's potential. The scores are $\hat\beta=\beta+e_\beta$ and $\hat\theta=\theta+e_\theta$, where the errors are jointly Gaussian, have mean zero and variances $\sigma^2_\beta(n)$ and $\sigma^2_\theta$, and are independent of $(\beta,\theta)$. Baseline prioritization ranks by $\hat\theta$ in the direction of its correlation with $\beta$.
\end{assumption}

Because probabilities and their differences are bounded, this Gaussian case is an approximation to our setting. The BATE is linear in the baseline here, ensuring signal monotonicity and making the bar \eqref{eq:bar} exact. We write $\mathcal U=(1-P)\Var(\beta)$ and $\Var(m_\theta)=P\Var(\beta)$ for the effect variance the baseline cannot and can see, and $\rho^2_\theta=\Var(\theta)/\{\Var(\theta)+\sigma^2_\theta\}$ for its recovery.

\begin{samepage}
\begin{proposition}[The noise an effect estimate can afford]\label{prop:noise}
Under \cref{ass:gaussian}, if $P>0$, $\hat\beta$ uniformly dominates $\hat\theta$ if and only if
\begin{equation}
  \sigma^2_\beta(n)\;\le\;\frac{\overbrace{\mathcal U}^{\text{never seen}}+\overbrace{\Var(m_\theta)\,(1-\rho^2_\theta)}^{\text{lost to noise}}}{P\,\rho^2_\theta}.
\end{equation}
If $P=0$, $\hat\beta$ uniformly dominates $\hat\theta$ at every finite noise variance.
\end{proposition}
\end{samepage}

This is the bar \eqref{eq:bar} expressed in terms of noise, using $\rho^2_\beta=\Var(\beta)/\{\Var(\beta)+\sigma^2_\beta\}$. The numerator captures what the baseline leaves unexplained: forgone heterogeneity plus variation lost to estimation noise. The more it leaves unexplained, the more noise an effect estimate can afford.

\begin{proof}[Proof of \cref{prop:noise}]
Orient the baseline in the direction of its correlation with $\beta$. For jointly Gaussian variables, $R^2_{\beta\mid S}=\Corr(\beta,S)^2$, and since the errors are independent of $(\beta,\theta)$ (though possibly correlated with each other),
\begin{equation}
\rho^2_\beta=R^2_{\beta\mid\hat\beta}=\frac{\Var(\beta)}{\Var(\beta)+\sigma^2_\beta},\qquad R^2_{\beta\mid\hat\theta}=\frac{\Cov(\beta,\theta)^2}{\Var(\beta)\{\Var(\theta)+\sigma^2_\theta\}}=P\rho^2_\theta.
\end{equation}
If $P>0$, both scores are positively correlated with $\beta$. Each expected effect $m_S$ is then linear and increasing in $S$, so both scores satisfy signal monotonicity, and the two expected effects are normal with the same mean $\E[\beta]$, so their CDFs cross at most once. \Cref{prop:bar}(b) therefore gives uniform dominance if and only if $\Var(\beta)/\{\Var(\beta)+\sigma^2_\beta\}\ge P\rho^2_\theta$, that is, $\sigma^2_\beta\le(1-P\rho^2_\theta)\Var(\beta)/(P\rho^2_\theta)$, where $(1-P\rho^2_\theta)\Var(\beta)=(1-P)\Var(\beta)+P(1-\rho^2_\theta)\Var(\beta)=\mathcal U+\Var(m_\theta)(1-\rho^2_\theta)$. If $P=0$, $\hat\theta$ is independent of $\beta$ and ranks at random, while $\hat\beta$ satisfies signal monotonicity, so it dominates.
\end{proof}

Converting the noise allowance to a sample size requires an estimator's noise rate. In the cell experiment, $n$ counts treated people.

\paragraph{Sampling variance in the cell experiment.}\phantomsection\label{app:noise}
With $n/K$ treated people in each cell, a cell's treated mean has variance $K\theta_1(1-\theta_1)/n$. With the baseline known, the cell estimator subtracts it from the treated mean, so its variance is $Kw/n$ with $w=\theta_1(1-\theta_1)$, and averaging over cells gives $\sigma^2_\beta(n)=K\bar w/n$. When the baseline must be estimated from the same experiment, with $n/K$ untreated people in each cell as well, a cell's untreated mean has variance $K\theta(1-\theta)/n$, and the cell estimator is the difference in means, with variance $K\{\theta_1(1-\theta_1)+\theta(1-\theta)\}/n$. Averaging over cells gives $K(\bar w_0+\bar w)/n$, with $\bar w_0=\E[\theta(1-\theta)]$; when effects are small, $\bar w\approx\bar w_0$ and this is about twice the variance of the untreated mean.

\begin{corollary}[Experiment size]\label{cor:size}
Under the Gaussian approximation with $\sigma^2_\beta(n)=K\bar w/n$, a known baseline, and $\mathcal U>0$, CATE prioritization uniformly dominates baseline prioritization if and only if
\begin{equation}
  n\;\ge\;n^*=\frac{K\bar w\,P}{\mathcal U}=\frac{K\bar w}{\Var(\beta)}\cdot\frac{P}{1-P}.
\end{equation}
\end{corollary}

\begin{proof}[Proof of \cref{cor:size}]
If $P=0$, the claim follows from \cref{prop:noise}. Otherwise, with a known baseline, $\rho^2_\theta=1$, and \cref{prop:noise} requires $K\bar w/n\le(1-P)\Var(\beta)/P=\mathcal U/P$, that is, $n\ge K\bar wP/\mathcal U$.
\end{proof}

\Cref{sec:thebar} discusses what moves $n^*$ and compares it with simulations.

\paragraph{Baselines from the same experiment.}\phantomsection\label{app:same}
If the baseline is estimated from an untreated arm of the same size as the treated one, then $\sigma^2_\theta=K\bar w_0/n$, the effect estimate is the difference in means, with $\sigma^2_\beta=K(\bar w_0+\bar w)/n$, and the condition of \cref{prop:noise} becomes $\Var(\beta)/\{\Var(\beta)+K(\bar w_0+\bar w)/n\}\ge P\Var(\theta)/\{\Var(\theta)+K\bar w_0/n\}$. Rearranging, CATE prioritization wins at every size if $P(\bar w_0+\bar w)\le\kappa\bar w_0$, and otherwise when
\begin{equation}\label{eq:nsame}
n\ge\frac{K}{\mathcal U}\big\{P(\bar w_0+\bar w)-\kappa\bar w_0\big\},\qquad\kappa=\frac{\Var(\beta)}{\Var(\theta)}.
\end{equation}
Compared with $n^*$, which uses the known baseline and so only the treated arm's noise, the requirement is $1+(\bar w_0/\bar w)(1-\kappa/P)$ times as large, about $2-\kappa/P$ when effects are small and $\bar w_0\approx\bar w$. Noise in the estimated baseline offsets only the term $\kappa/P$, where $\kappa$, the variance of effects relative to that of the baseline, is small under modest shifts. In the example of \cref{sec:thebar}, if baselines have a standard deviation of 5 percentage points, then $\kappa=0.16$; taking $\bar w_0\approx\bar w$, the requirement rises by a factor of $1.8$ when a fifth of effect variance is forgone. The correlation between the two estimates' errors does not matter, since each score is compared with $\beta$ separately.

\Cref{app:learned} compares this prediction with simulations.

\paragraph{Beyond the Gaussian case.}\phantomsection\label{app:sufficient}
Suppose the baseline is known, and let $\hat\beta=\beta+e_\beta$ with $\E[e_\beta\mid\beta]=0$ and $\E[e_\beta^2]=\sigma^2_\beta$. The conditional mean $\E[\beta\mid\hat\beta]$ predicts $\beta$ at least as well as the best linear predictor, so $R^2_{\beta\mid\hat\beta}\ge\Corr(\beta,\hat\beta)^2=\Var(\beta)/\{\Var(\beta)+\sigma^2_\beta\}$, while $R^2_{\beta\mid m_\theta}=P$. If $\hat\beta$ and $m_\theta$ meet the conditions of \cref{prop:bar}(b), then $\Var(\beta)/\{\Var(\beta)+\sigma^2_\beta\}\ge P$ suffices for $\hat\beta$ to uniformly dominate BATE prioritization, and with $\sigma^2_\beta=K\bar w/n$ this is $n\ge n^*$. The condition need not be necessary, because the first inequality can be strict.

In the running example, winning at every capacity takes longer than $n^*$ (last column of \cref{tab:fraction}), so the conditions under which $n^*$ suffices do not hold there. What $n^*$ tracks is the size at which CATE prioritization wins in normalized Qini (\cref{tab:avg}); \cref{app:simdetails} reports its accuracy.

% =====================================================================
\section{Simulation Details}\label{app:simdetails}

\paragraph{Populations.} Noise is standard logistic and the expected shift is $\bar\delta=0.5$, so under a common shift effects peak at $F(-0.25)\approx44\%$. Each population spreads systematic indices evenly over an interval: from $F^{-1}(0.03)$ to $F^{-1}(0.20)$ (rare), from $F^{-1}(0.60)$ to $F^{-1}(0.95)$ (common), and from $F^{-1}(0.10)$ to $F^{-1}(0.80)$ (straddling). The simulated populations have $K=50$ equal-mass cells in 10 baseline groups of 5, with group indices evenly spaced over the interval, endpoints included. Within each group, the five cells have shifts $0.5+\eta$, where $\eta$ is evenly spaced around zero with standard deviation $0.5r$ ($\eta=0$ under a common shift). The straddling population uses $r=0$ or $r=0.4$; the $\delta=2$ example in \cref{sec:locate} uses the same cells with a common shift of 2. The reversed population (last row of \cref{tab:fraction}) is the rare population with the group standard deviation of $\eta$ proportional to $\{\max_h f(\mu_h+0.5)/f(\mu_g+0.5)\}^{1.5}$ and averaging $0.2$, as with $r=0.4$. It falls from 0.53 in the lowest-baseline group to 0.056 in the highest, making the sensitivity-weighted spread about twice as large at the bottom of the ranking as at the top.

\paragraph{Rankings.} In the straddling population, baseline prioritization ranks in the direction that performs better on average (from the highest baseline down when $\delta=0.5$), threshold prioritization ranks cells by $-|\theta_c-1/2|$, and BATE prioritization ranks them by their group's mean effect. Estimated BATE prioritization uses the one-parameter logit, which solves $\sum_c\{\bar Y_{1c}-F(\mu_c+\hat\delta)\}=0$ over the treated arm with the baseline known and ranks cells by $-|\mu_c+\hat\delta/2|$; under heterogeneous shifts it can recover the BATE ordering without recovering its levels. In the rare, common, and reversed populations, group mean effects are monotone in the baseline for every $r$, so baseline prioritization in the matching direction coincides with BATE prioritization; cells in a group tie and contribute the group's mean effect.

\paragraph{Experiments.} Each experiment assigns $m$ treated people to every cell ($n=50m$ treated people in total) and draws binomial outcomes. We use 120 log-spaced grid values from 300 to 30 million, set $m$ to the grid value divided by 100 and rounded down (giving $n$ from 150 to 15.8 million treated people), run 3{,}000 replications at each size, and pool grid values that give the same $n$. With a known baseline, the cell estimator ranks cells by $\bar Y_{1c}-\theta_c$. When baselines come from the same experiment, each cell also gets $m$ untreated people; the cell estimator ranks by $\bar Y_{1c}-\bar Y_{0c}$ and baseline prioritization ranks by $\bar Y_{0c}$. Ties are broken at random. The pooled-estimator comparisons in \cref{app:pooled} use a separate 3{,}000-replication run on the same grid in which all estimators share the same draws.

\paragraph{Evaluation and crossovers.} At each size, we average the 3{,}000 prioritization curves at $u=j/50$ for $j=1,\ldots,50$. Normalized Qini \eqref{eq:qini-share} takes the discrete average of each curve over these 50 fractions (the discrete counterpart of \eqref{eq:average-value}, under which random targeting averages $0.51\,\E[\beta]$ rather than $\E[\beta]/2$). A reported crossover at fraction $u=j/50$, or in normalized Qini, is the first tested size from which CATE prioritization's average curve matches or exceeds the baseline curve at that size and every larger tested size. Because these are crossings of averaged curves on a finite grid, they locate each threshold only up to the grid spacing and say nothing about any single experiment.

\paragraph{Accuracy of the approximations.} The formula $n^*$ computes $P=\Var(m_\theta)/\{\Var(m_\theta)+\mathcal U\}$ from the 10-group structure. With a fixed number of cells, misranking probabilities eventually fall exponentially in $n$, so the simulations test the formulas as approximations over practical experiment sizes rather than as limits. For the known-baseline scenarios in \cref{tab:avg}, $n^*$ is between 0.76 and 1.25 times the observed crossover; in the reversed population, $n^*\approx11{,}900$ against an observed crossover of 10{,}100 treated people. \Cref{thm:margin} and \cref{cor:narrow} assume small unpredictable shifts, small estimation noise, and treatment fractions away from the extremes. The cell simulations relax the first two---in \cref{tab:fraction}, the experiment needed still rises with $u$ even at $r=0.8$---while the continuous population below checks \cref{thm:margin} directly.

\paragraph{Continuous population.} To check \cref{thm:margin} without cells, we draw 400{,}000 people with indices uniform on $[F^{-1}(0.03),F^{-1}(0.20)]$, expected shift $0.5$, and $\eta\sim N(0,(0.5r)^2)$ for $r=0.1$, 0.05, and 0.025. The effect estimate is $\hat\beta=\beta+\sigma\sqrt{w(X)}\,e$ with $e\sim N(0,1)$, averaged over four draws of $e$. At each $u\in\{0.05,0.10,\dots,0.95\}$, bisection finds the $\sigma^2$ at which CATE and baseline prioritization tie, which we compare against the prediction $\Var(\beta\mid\mu_u)/\E[w\mid\mu_u]$ computed by quadrature. Over fractions from 20\% to 80\%, the ratio of predicted to simulated $\sigma^2$ ranges from 0.87 to 1.02, 0.98 to 1.03, and 0.97 to 1.04 for the three values of $r$. The simulated experiment needed rises with $u$ through 80\%, as \cref{cor:narrow} predicts; at the narrowest fractions it is even smaller than predicted, because the edge of the effect distribution limits how many people noise can swap.

% \paragraph{Blending with the BATE.} To check \cref{cor:shrink}, we use the same continuous population with $r=0.1$ and $0.05$ and estimates $\hat\beta_\omega=m_\theta+\omega\{\beta-m_\theta+\sigma\sqrt{w(X)}\,e\}$, with weights $\omega=1$, $0.5$, and $0.25$, and $\omega$ drawn uniformly from $[0,0.6]$ in each of four replications. We set $\sigma^2$ so that forgone heterogeneity at the median baseline is twice, half, and a fifth of the noise there, and compare the simulated gain over BATE prioritization at $u=0.2$, 0.5, and 0.8 with \eqref{eq:shrink}, averaged over $\omega$ when it is random. The predicted and simulated gains have the same sign in all 72 comparisons. With $r=0.05$, they agree within 7\% wherever the gain exceeds $10^{-5}$ in absolute value, and smaller gains differ by at most $1.2\times10^{-6}$. With $r=0.1$, the agreement is looser when the noise is five times forgone heterogeneity, as expected when errors are no longer small. \Cref{fig:shrink} uses the case with $r=0.05$ and forgone heterogeneity at the median half the noise, with fixed weights from 0.1 to 1 and eight noise draws, and divides each gain by the simulated gain of ranking by the true CATE, $V(u;\beta)-V(u;m_\theta)$.

\paragraph{Code.} The \texttt{code} folder reproduces every simulated number, table, and figure, with fixed random seeds. \texttt{run\_simulations.py}, \texttt{run\_pooled.py}, and \texttt{run\_continuous.py} run the experiments, \texttt{run\_equal.py} and \texttt{run\_equal\_pooled.py} the experiments with a learned baseline; \texttt{make\_outputs.py}, \texttt{make\_outputs\_pooled.py}, and \texttt{make\_outputs\_equal.py} print the tables, the numbers quoted in the text, and the figure coordinates from the saved results, with sizes in treated people, or in all observations when the baseline must be learned. Its README maps each result to the script that produces it.

\section{Robustness of the Main Results}\label{app:robust}

The main analysis makes two simplifications that could tilt the comparison between rankings. It uses a simple effect estimator, the cell estimator, and it treats the baseline as known. This appendix relaxes each in turn. The mechanisms and qualitative conclusions survive both.

\subsection{A Better Effect Estimator}\label{app:pooled}

The cell estimator treats every estimated difference between cells as signal, so \cref{sec:size} uses it as a simple benchmark. A better estimator changes how much data CATE prioritization needs, but not the paper's conclusions. The \textbf{pooled estimator} shrinks estimated cell shifts toward the one-parameter logit's common shift by empirical Bayes \citep{morris1983parametric}, retaining more variation when the cells differ more than noise alone would explain. It nearly matches estimated BATE prioritization when the baseline misses nothing. It needs 25\% to 55\% less data than the cell estimator when the baseline misses about half of effect variance or less, and about a fifth more when it misses most of it. Narrow campaigns still need less, and the reversed population still reverses (\cref{tab:pooled}).

\paragraph{Estimator.} The pooled estimator shrinks on the log-odds scale and by a different amount in each cell, depending on that cell's noise. With $m$ treated people per cell and the baseline known, we compute the treated mean $\hat\theta_{1c}=(\bar Y_{1c}m+1/2)/(m+1)$, the cell's log-odds shift $\hat\delta_c=F^{-1}(\hat\theta_{1c})-\mu_c$, and its delta-method variance $v_c=1/\{m\hat\theta_{1c}(1-\hat\theta_{1c})\}$. With $\hat\delta$ the one-parameter logit's common shift, the variance of true shifts across cells is estimated by moments, $\hat\tau^2=\max\{0,K^{-1}\sum_c[(\hat\delta_c-\hat\delta)^2-v_c]\}$, and each shift is shrunk toward $\hat\delta$ by its estimated signal share, $\tilde\delta_c=\hat\delta+\hat\tau^2(\hat\delta_c-\hat\delta)/(\hat\tau^2+v_c)$, so noisier cells shrink more strongly. Cells are ranked by $F(\mu_c+\tilde\delta_c)-\theta_c$. Because the pooled estimator ranks exactly like the logit when $\hat\tau^2=0$, and so ties with baseline prioritization in the rare, common, and reversed populations, we compare it at a given fraction by the size from which it beats baseline prioritization by at least 1\% of the best ranking's value, rather than by the size from which it matches it.

\paragraph{Results.} In the straddling population, under a common shift, where there is nothing to find, the pooled estimator's estimated shift variance is exactly zero in 55\% to 65\% of experiments, so it ranks like the logit; its normalized Qini dips to about 95\% around 1{,}600 treated people and stays above 99\% from about 12{,}000, whereas the cell estimator needs about 650{,}000. When shifts also vary within baseline groups ($r=0.4$), it overtakes estimated BATE prioritization at about 350 treated people, against 1{,}600 for the cell estimator. \Cref{tab:pooled} reports the rare, common, and reversed populations.

\begin{table}[t]
\centering
\caption{\textbf{The pooled estimator in the running example.} Smallest experiment size (number of treated people) at which each estimator consistently closes half the gap in normalized Qini between BATE prioritization and the best ranking, and at which the pooled estimator consistently beats baseline prioritization by at least 1\% of the best ranking's value among the top 10\% and top 80\%, in simulations of the running example. The baseline is known. A dash means no such win up to 15 million.}
\label{tab:pooled}
\vspace{4pt}
\small
\begin{tabular}{lcrrrrr}
\toprule
& & & \multicolumn{2}{c}{Half the gap} & \multicolumn{2}{c}{Pooled, 1\% better} \\
\cmidrule(lr){4-5}\cmidrule(l){6-7}
Population & $r$ & Forgone share & Cell & Pooled & Top 10\% & Top 80\% \\
\midrule
Rare   & 0.8 & 81\% & 2{,}150 & 2{,}600 & 900 & 1{,}600 \\
Rare   & 0.4 & 53\% & 13{,}500 & 10{,}100 & 1{,}450 & 12{,}250 \\
Rare   & 0.2 & 22\% & 70{,}150 & 39{,}250 & 6{,}850 & -- \\
Rare   & 0.1 & 7\% & 271{,}800 & 125{,}350 & 77{,}250 & -- \\
Common & 0.8 & 72\% & 2{,}600 & 3{,}150 & 1{,}050 & 2{,}150 \\
Common & 0.4 & 37\% & 18{,}100 & 11{,}150 & 2{,}150 & 203{,}300 \\
Common & 0.2 & 13\% & 70{,}150 & 39{,}250 & 8{,}350 & -- \\
Common & 0.1 & 4\% & 246{,}700 & 113{,}750 & 103{,}300 & -- \\
\midrule
Reversed &  & 38\% & 26{,}650 & 18{,}100 & 47{,}600 & 2{,}600 \\
\bottomrule
\end{tabular}
\end{table}

\subsection{Learning the Baseline from Data}\label{app:learned}

The main text treats the baseline as known, as when it is learned from abundant untreated data. We relax this in two ways.

\paragraph{Baselines from the same experiment.} Suppose the experiment includes an untreated arm as large as the treated one, which provides the baseline. The cell estimator must then use a difference in means, roughly doubling its noise, while baseline prioritization is barely affected (\cref{sec:easier}). \Cref{tab:same} shows the result. Estimating the baseline from the experiment raises the size CATE prioritization needs to win in normalized Qini by 20\% to 140\% in seven of the eight scenarios. The exception is the rare population with 81\% forgone, where the baseline is estimated so poorly that CATE prioritization wins at the smallest size simulated. The Gaussian prediction, extended to this design in \cref{app:same}, stays within about 30\% of these sizes, slightly less close than $n^*$ with the baseline known.

\begin{table}[t]
\centering
\caption{\textbf{When the baseline comes from the same experiment.} Crossovers (defined as in \cref{tab:avg}), in treated people, for CATE prioritization to match or exceed the normalized Qini of baseline prioritization in the running example, when the baseline is estimated from an untreated arm as large as the treated one: observed in simulations and predicted by the Gaussian formula of \cref{app:same}. The known-baseline column repeats \cref{tab:avg}. ``Any'' means the formula predicts a win at every size; 150 is the smallest size simulated.}
\label{tab:same}
\vspace{4pt}
\small
\begin{tabular}{lcrrrr}
\toprule
& & & Baseline known & \multicolumn{2}{c}{Same experiment} \\
\cmidrule(l){5-6}
Population & $r$ & Forgone share & Observed & Predicted & Observed \\
\midrule
Rare   & 0.8 & 81\% & 550 & any & 150 \\
Rare   & 0.4 & 53\% & 5{,}100 & 4{,}400 & 6{,}200 \\
Rare   & 0.2 & 22\% & 32{,}350 & 40{,}600 & 52{,}450 \\
Rare   & 0.1 & 7\% & 125{,}350 & 204{,}500 & 223{,}950 \\
Common & 0.8 & 72\% & 700 & 740 & 950 \\
Common & 0.4 & 37\% & 7{,}550 & 12{,}100 & 16{,}400 \\
Common & 0.2 & 13\% & 35{,}600 & 71{,}300 & 85{,}100 \\
Common & 0.1 & 4\% & 113{,}750 & 319{,}000 & 271{,}800 \\
\bottomrule
\end{tabular}
\end{table}

\paragraph{Equal observations for every ranking.} A stricter comparison gives every ranking the same total number of observations $n$: untreated people for baseline and threshold prioritization, and an experiment for BATE and CATE prioritization. The mechanisms survive, but their costs move.

Rankings by the baseline itself barely notice. With 50 cells, baseline prioritization comes within 1\% of its value under a known baseline from about 5{,}000 untreated people in the common population and 12{,}500 in the rare one, because the same noise disturbs baselines less than effects (\cref{sec:easier}). Learning the BATE changes more. In the straddling population with a common shift, estimated BATE prioritization reaches 89\% of the best normalized Qini with 3{,}000 observations and 99\% with 30{,}000, against 99\% with 150 treated people when the baseline is known (\cref{fig:learned}). Nearly all of the difference comes from estimating each cell's baseline, not the shift: in a balanced experiment of 1{,}000, estimating only the baseline leaves a normalized Qini of 58\%, the same as estimating both, while estimating only the shift leaves 99.6\%. Fixing the ordering remains cheap; what takes data is learning the baseline it is applied to.

Because each cell's baseline is learned from its own untreated people while the shift is learned from everyone treated, an experiment for BATE prioritization should be mostly untreated. A leading-order calculation, treating the misplacement of the selected interval like the displacement in \cref{prop:secondorder}, puts the loss-minimizing treated share at $1/\{1+(4K-3)^{1/2}\}$, or 6.6\% with $K=50$; with 10 baseline groups of 2, 5, or 20 cells, simulations find the loss smallest near the predicted 10.2\%, 6.6\%, and 3.4\%, and a balanced split raises it by 1.7 to 2.0 times.

Given such an experiment for BATE prioritization and a balanced one for the cell estimator, the main conclusions still hold. CATE prioritization still overtakes estimated BATE prioritization when shifts vary within baseline groups, at about 5{,}000 observations in the straddling population (\cref{fig:learned}B), and narrow campaigns still need less: in the rare population with 53\% forgone, it beats baseline prioritization among the top 10\% with 2{,}400 observations but needs 36{,}200 among the top 80\%. The pooled estimator keeps its advantage over the cell estimator: in the straddling population ($r=0.4$), it overtakes estimated BATE prioritization at about 2{,}400 observations, and when 53\% or less of effect variance is forgone, it closes half the gap between BATE prioritization and the best ranking with 50\% to 75\% of the observations the cell estimator needs. Under a common shift, it trails estimated BATE prioritization, because it learns each cell's baseline from half as many people.

\paragraph{Design.} With 50 cells, a budget of $n=100m$ gives $2m$ observations per cell. Baseline and threshold prioritization use all $2m$ untreated people; CATE prioritization (cell estimator) uses $m$ untreated and $m$ treated people and ranks by difference in means; estimated BATE prioritization uses $\max\{1,\mathrm{round}(0.066\cdot2m)\}$ treated people and the remaining $m_0$ untreated. It estimates each cell's baseline log-odds from its untreated mean with a continuity correction, $F^{-1}\{(\bar Y_{0c}m_0+1/2)/(m_0+1)\}$, and the common shift from its treated people with these estimates held fixed. We draw the samples jointly from a shared pool of $2m$ untreated and $m$ treated people per cell, which leaves each ranking's sampling distribution unchanged and makes comparisons between rankings less noisy; drawing them independently leaves the crossovers unchanged. Averages use 3{,}000 replications. In a separate run, also with 3{,}000 replications, the pooled estimator uses the same balanced experiment as the cell estimator, estimates each cell's log-odds shift by the difference of the continuity-corrected arm log-odds, and shrinks it as in \cref{app:pooled} toward the one-parameter logit's common shift, fitted with the control-arm baselines. The code is \texttt{run\_equal.py --own}, \texttt{alloc\_theory\_check.py}, and \texttt{robustness.py} in the \texttt{code} folder.

\begin{figure}[t]
\centering
\begin{minipage}[t]{0.49\textwidth}
\centering
\begin{tikzpicture}
\begin{axis}[
    width=\linewidth, height=6cm,
    title={\small (A) Common shift},
    xmode=log, xmin=300, xmax=3.2e7, ymin=0, ymax=102,
    xlabel={Observations $n$}, ylabel={Normalized Qini (\%)},
    xtick={1e3,1e4,1e5,1e6,1e7}, xticklabels={$10^3$,$10^4$,$10^5$,$10^6$,$10^7$},
    legend columns=4, legend to name=learnedleg, legend style={font=\scriptsize, draw=none, /tikz/every even column/.append style={column sep=10pt}},
    legend cell align=left, grid=major, grid style={gray!15},
    every axis plot/.append style={line width=1.3pt}, tick label style={font=\scriptsize}, label style={font=\small}
]
\addplot[orange!80!black, densely dashed] coordinates {(300,38.29) (400,47.51) (500,54.04) (600,59.22) (700,63.21) (800,66.36) (900,69.06) (1000,71.55) (1100,73.37) (1200,75.26) (1300,76.79) (1400,78.29) (1600,80.54) (1800,82.26) (1900,83.09) (2100,84.46) (2400,86.20) (2600,87.14) (2900,88.46) (3200,89.42) (3500,90.36) (3900,91.24) (4300,92.02) (4700,92.72) (5200,93.41) (5700,93.96) (6300,94.57) (6900,95.11) (7700,95.66) (8400,96.05) (9300,96.53) (10200,96.83) (11300,97.19) (12400,97.47) (13700,97.79) (15100,98.02) (16700,98.24) (18400,98.47) (20200,98.62) (22300,98.78) (24500,98.93) (27000,99.04) (29800,99.13) (32800,99.22) (36200,99.30) (39800,99.35) (43900,99.42) (48400,99.47) (53300,99.50) (58700,99.54) (64700,99.58) (71200,99.60) (78500,99.64) (86400,99.67) (95200,99.69) (104900,99.71) (115600,99.74) (127300,99.76) (140300,99.79) (154500,99.80) (170200,99.82) (187500,99.85) (206600,99.86) (227500,99.89) (250700,99.90) (276100,99.92) (304200,99.93) (335100,99.95) (369100,99.96) (406600,99.97) (447900,99.98) (493400,99.98) (543600,99.98) (598800,99.99) (659600,99.99) (726600,100.00) (800500,100.00) (881800,100.00) (971300,100.00) (1070000,100.00) (1178700,100.00) (1298500,100.00) (1430400,100.00) (1575700,100.00) (1735700,100.00) (1912100,100.00) (2106300,100.00) (2320300,100.00) (2556000,100.00) (2815600,100.00) (3101600,100.00) (3416700,100.00) (3763800,100.00) (4146100,100.00) (4567300,100.00) (5031300,100.00) (5542400,100.00) (6105400,100.00) (6725600,100.00) (7408800,100.00) (8161400,100.00) (8990400,100.00) (9903700,100.00) (10909700,100.00) (12017900,100.00) (13238800,100.00) (14583600,100.00) (16065000,100.00) (17696900,100.00) (19494600,100.00) (21474900,100.00) (23656400,100.00) (26059500,100.00) (28706600,100.00) (31622700,100.00)};
\addlegendentry{Estimated BATE prioritization}
\addplot[blue!70!black, solid] coordinates {(300,6.53) (400,7.62) (500,8.46) (600,9.02) (700,9.73) (800,10.82) (900,11.46) (1000,11.93) (1100,12.57) (1200,12.94) (1300,14.07) (1400,14.37) (1600,14.92) (1800,16.53) (1900,16.45) (2100,17.35) (2400,18.54) (2600,19.19) (2900,19.83) (3200,20.56) (3500,21.73) (3900,23.62) (4300,24.75) (4700,25.71) (5200,26.46) (5700,28.10) (6300,29.84) (6900,30.59) (7700,31.70) (8400,33.30) (9300,34.48) (10200,36.44) (11300,38.03) (12400,39.35) (13700,41.06) (15100,42.72) (16700,44.85) (18400,46.31) (20200,48.36) (22300,50.52) (24500,52.03) (27000,54.01) (29800,55.86) (32800,58.05) (36200,60.21) (39800,62.05) (43900,64.10) (48400,66.20) (53300,68.25) (58700,69.54) (64700,71.39) (71200,73.22) (78500,75.01) (86400,76.70) (95200,78.19) (104900,79.68) (115600,81.23) (127300,82.49) (140300,83.84) (154500,84.93) (170200,86.25) (187500,87.27) (206600,88.33) (227500,89.33) (250700,90.15) (276100,90.98) (304200,91.77) (335100,92.49) (369100,93.11) (406600,93.72) (447900,94.29) (493400,94.84) (543600,95.27) (598800,95.70) (659600,96.15) (726600,96.45) (800500,96.82) (881800,97.11) (971300,97.39) (1070000,97.65) (1178700,97.87) (1298500,98.08) (1430400,98.27) (1575700,98.40) (1735700,98.56) (1912100,98.69) (2106300,98.80) (2320300,98.91) (2556000,99.00) (2815600,99.08) (3101600,99.15) (3416700,99.20) (3763800,99.27) (4146100,99.32) (4567300,99.38) (5031300,99.42) (5542400,99.47) (6105400,99.51) (6725600,99.55) (7408800,99.59) (8161400,99.62) (8990400,99.66) (9903700,99.69) (10909700,99.73) (12017900,99.75) (13238800,99.78) (14583600,99.80) (16065000,99.83) (17696900,99.85) (19494600,99.87) (21474900,99.89) (23656400,99.90) (26059500,99.92) (28706600,99.93) (31622700,99.94)};
\addlegendentry{CATE prioritization (cell)}
\addplot[green!55!black, densely dashdotted] coordinates {(300,44.30) (400,51.96) (500,57.23) (600,61.63) (700,65.11) (800,67.73) (900,70.01) (1000,72.01) (1100,73.51) (1200,74.93) (1300,76.15) (1400,77.32) (1600,78.96) (1800,80.43) (1900,80.94) (2100,82.12) (2400,83.45) (2600,84.10) (2900,85.04) (3200,85.76) (3500,86.51) (3900,87.10) (4300,87.69) (4700,88.26) (5200,88.79) (5700,89.21) (6300,89.60) (6900,89.96) (7700,90.38) (8400,90.62) (9300,90.92) (10200,91.17) (11300,91.39) (12400,91.59) (13700,91.78) (15100,91.86) (16700,92.00) (18400,92.07) (20200,92.19) (22300,92.26) (24500,92.28) (27000,92.31) (29800,92.33) (32800,92.34) (36200,92.38) (39800,92.32) (43900,92.28) (48400,92.30) (53300,92.28) (58700,92.21) (64700,92.21) (71200,92.14) (78500,92.16) (86400,92.09) (95200,92.06) (104900,91.99) (115600,92.00) (127300,91.91) (140300,91.91) (154500,91.83) (170200,91.78) (187500,91.74) (206600,91.68) (227500,91.66) (250700,91.59) (276100,91.51) (304200,91.47) (335100,91.39) (369100,91.34) (406600,91.26) (447900,91.15) (493400,91.12) (543600,91.01) (598800,90.93) (659600,90.85) (726600,90.73) (800500,90.67) (881800,90.59) (971300,90.48) (1070000,90.35) (1178700,90.28) (1298500,90.14) (1430400,90.05) (1575700,89.93) (1735700,89.83) (1912100,89.71) (2106300,89.60) (2320300,89.49) (2556000,89.36) (2815600,89.22) (3101600,89.11) (3416700,89.02) (3763800,88.87) (4146100,88.79) (4567300,88.68) (5031300,88.58) (5542400,88.47) (6105400,88.40) (6725600,88.30) (7408800,88.23) (8161400,88.16) (8990400,88.10) (9903700,88.04) (10909700,87.99) (12017900,87.96) (13238800,87.92) (14583600,87.90) (16065000,87.87) (17696900,87.85) (19494600,87.84) (21474900,87.83) (23656400,87.83) (26059500,87.82) (28706600,87.82) (31622700,87.82)};
\addlegendentry{Threshold prioritization}
\addplot[gray!60, dashdotted] coordinates {(300,17.61) (400,18.97) (500,19.90) (600,20.53) (700,21.06) (800,21.46) (900,21.68) (1000,21.99) (1100,22.23) (1200,22.39) (1300,22.60) (1400,22.75) (1600,22.97) (1800,23.11) (1900,23.29) (2100,23.50) (2400,23.67) (2600,23.73) (2900,23.87) (3200,23.98) (3500,24.07) (3900,24.17) (4300,24.25) (4700,24.37) (5200,24.43) (5700,24.53) (6300,24.57) (6900,24.63) (7700,24.67) (8400,24.73) (9300,24.80) (10200,24.83) (11300,24.87) (12400,24.91) (13700,24.95) (15100,24.98) (16700,25.01) (18400,25.05) (20200,25.07) (22300,25.09) (24500,25.12) (27000,25.14) (29800,25.16) (32800,25.17) (36200,25.18) (39800,25.18) (43900,25.19) (48400,25.19) (53300,25.20) (58700,25.20) (64700,25.20) (71200,25.20) (78500,25.20) (86400,25.20) (95200,25.20) (104900,25.20) (115600,25.20) (127300,25.20) (140300,25.20) (154500,25.20) (170200,25.20) (187500,25.20) (206600,25.20) (227500,25.20) (250700,25.20) (276100,25.20) (304200,25.20) (335100,25.20) (369100,25.20) (406600,25.20) (447900,25.20) (493400,25.20) (543600,25.20) (598800,25.20) (659600,25.20) (726600,25.20) (800500,25.20) (881800,25.20) (971300,25.20) (1070000,25.20) (1178700,25.20) (1298500,25.20) (1430400,25.20) (1575700,25.20) (1735700,25.20) (1912100,25.20) (2106300,25.20) (2320300,25.20) (2556000,25.20) (2815600,25.20) (3101600,25.20) (3416700,25.20) (3763800,25.20) (4146100,25.20) (4567300,25.20) (5031300,25.20) (5542400,25.20) (6105400,25.20) (6725600,25.20) (7408800,25.20) (8161400,25.20) (8990400,25.20) (9903700,25.20) (10909700,25.20) (12017900,25.20) (13238800,25.20) (14583600,25.20) (16065000,25.20) (17696900,25.20) (19494600,25.20) (21474900,25.20) (23656400,25.20) (26059500,25.20) (28706600,25.20) (31622700,25.20)};
\addlegendentry{Baseline prioritization}
\end{axis}
\end{tikzpicture}
\end{minipage}\hfill
\begin{minipage}[t]{0.49\textwidth}
\centering
\begin{tikzpicture}
\begin{axis}[
    width=\linewidth, height=6cm,
    title={\small (B) 75\% of effect variance forgone},
    xmode=log, xmin=300, xmax=3.2e7, ymin=0, ymax=102,
    xlabel={Observations $n$}, ylabel={Normalized Qini (\%)},
    xtick={1e3,1e4,1e5,1e6,1e7}, xticklabels={$10^3$,$10^4$,$10^5$,$10^6$,$10^7$},
    grid=major, grid style={gray!15},
    every axis plot/.append style={line width=1.3pt}, tick label style={font=\scriptsize}, label style={font=\small}
]
\addplot[orange!80!black, densely dashed] coordinates {(300,19.10) (400,23.46) (500,26.89) (600,29.35) (700,31.32) (800,33.11) (900,34.47) (1000,35.23) (1100,36.36) (1200,37.19) (1300,38.29) (1400,38.91) (1600,39.89) (1800,40.76) (1900,41.19) (2100,41.86) (2400,42.89) (2600,43.24) (2900,43.70) (3200,44.23) (3500,44.93) (3900,45.21) (4300,45.59) (4700,46.02) (5200,46.31) (5700,46.57) (6300,46.94) (6900,47.17) (7700,47.43) (8400,47.66) (9300,47.71) (10200,48.13) (11300,48.11) (12400,48.22) (13700,48.32) (15100,48.65) (16700,48.77) (18400,48.80) (20200,48.86) (22300,48.91) (24500,48.99) (27000,49.09) (29800,49.10) (32800,49.17) (36200,49.25) (39800,49.15) (43900,49.25) (48400,49.31) (53300,49.30) (58700,49.34) (64700,49.38) (71200,49.40) (78500,49.44) (86400,49.48) (95200,49.48) (104900,49.49) (115600,49.54) (127300,49.62) (140300,49.64) (154500,49.62) (170200,49.70) (187500,49.59) (206600,49.67) (227500,49.67) (250700,49.71) (276100,49.77) (304200,49.77) (335100,49.74) (369100,49.73) (406600,49.75) (447900,49.80) (493400,49.76) (543600,49.83) (598800,49.76) (659600,49.77) (726600,49.75) (800500,49.79) (881800,49.79) (971300,49.80) (1070000,49.78) (1178700,49.83) (1298500,49.79) (1430400,49.73) (1575700,49.84) (1735700,49.83) (1912100,49.77) (2106300,49.77) (2320300,49.79) (2556000,49.84) (2815600,49.78) (3101600,49.79) (3416700,49.81) (3763800,49.80) (4146100,49.78) (4567300,49.79) (5031300,49.84) (5542400,49.80) (6105400,49.82) (6725600,49.80) (7408800,49.80) (8161400,49.81) (8990400,49.81) (9903700,49.75) (10909700,49.83) (12017900,49.83) (13238800,49.77) (14583600,49.74) (16065000,49.83) (17696900,49.82) (19494600,49.84) (21474900,49.85) (23656400,49.77) (26059500,49.84) (28706600,49.78) (31622700,49.84)};
\addplot[blue!70!black, solid] coordinates {(300,12.20) (400,14.15) (500,15.67) (600,17.43) (700,18.26) (800,19.93) (900,20.97) (1000,22.72) (1100,23.23) (1200,24.59) (1300,24.97) (1400,26.12) (1600,28.08) (1800,29.61) (1900,30.38) (2100,32.02) (2400,34.08) (2600,34.76) (2900,36.60) (3200,38.02) (3500,39.76) (3900,41.70) (4300,43.51) (4700,45.16) (5200,46.84) (5700,48.55) (6300,50.39) (6900,51.98) (7700,54.27) (8400,56.07) (9300,58.18) (10200,59.89) (11300,61.98) (12400,64.04) (13700,65.48) (15100,67.61) (16700,69.25) (18400,71.23) (20200,72.93) (22300,74.60) (24500,76.24) (27000,77.86) (29800,79.40) (32800,80.75) (36200,82.02) (39800,83.31) (43900,84.72) (48400,85.79) (53300,86.92) (58700,87.88) (64700,88.91) (71200,89.66) (78500,90.52) (86400,91.33) (95200,92.08) (104900,92.66) (115600,93.35) (127300,93.89) (140300,94.40) (154500,94.93) (170200,95.36) (187500,95.76) (206600,96.15) (227500,96.53) (250700,96.82) (276100,97.11) (304200,97.39) (335100,97.62) (369100,97.83) (406600,98.02) (447900,98.23) (493400,98.38) (543600,98.54) (598800,98.68) (659600,98.78) (726600,98.89) (800500,99.00) (881800,99.09) (971300,99.18) (1070000,99.26) (1178700,99.32) (1298500,99.39) (1430400,99.44) (1575700,99.49) (1735700,99.54) (1912100,99.58) (2106300,99.62) (2320300,99.65) (2556000,99.69) (2815600,99.71) (3101600,99.74) (3416700,99.76) (3763800,99.78) (4146100,99.80) (4567300,99.82) (5031300,99.83) (5542400,99.85) (6105400,99.86) (6725600,99.87) (7408800,99.88) (8161400,99.89) (8990400,99.90) (9903700,99.91) (10909700,99.92) (12017900,99.92) (13238800,99.93) (14583600,99.94) (16065000,99.94) (17696900,99.95) (19494600,99.95) (21474900,99.96) (23656400,99.96) (26059500,99.96) (28706600,99.97) (31622700,99.97)};
\addplot[green!55!black, densely dashdotted] coordinates {(300,21.37) (400,25.03) (500,27.74) (600,29.86) (700,31.30) (800,32.81) (900,34.20) (1000,34.64) (1100,35.40) (1200,36.14) (1300,36.91) (1400,37.33) (1600,38.17) (1800,38.79) (1900,39.18) (2100,39.62) (2400,40.42) (2600,40.58) (2900,40.95) (3200,41.39) (3500,41.90) (3900,42.15) (4300,42.27) (4700,42.67) (5200,42.90) (5700,43.09) (6300,43.40) (6900,43.53) (7700,43.62) (8400,43.77) (9300,43.83) (10200,44.20) (11300,44.10) (12400,44.15) (13700,44.20) (15100,44.42) (16700,44.61) (18400,44.52) (20200,44.53) (22300,44.55) (24500,44.61) (27000,44.60) (29800,44.61) (32800,44.60) (36200,44.65) (39800,44.48) (43900,44.56) (48400,44.52) (53300,44.51) (58700,44.47) (64700,44.45) (71200,44.55) (78500,44.50) (86400,44.44) (95200,44.47) (104900,44.39) (115600,44.42) (127300,44.34) (140300,44.36) (154500,44.32) (170200,44.42) (187500,44.28) (206600,44.21) (227500,44.17) (250700,44.24) (276100,44.17) (304200,44.12) (335100,43.97) (369100,43.99) (406600,43.93) (447900,43.98) (493400,43.86) (543600,43.86) (598800,43.79) (659600,43.64) (726600,43.64) (800500,43.68) (881800,43.57) (971300,43.52) (1070000,43.43) (1178700,43.36) (1298500,43.29) (1430400,43.21) (1575700,43.29) (1735700,43.20) (1912100,43.09) (2106300,42.97) (2320300,42.98) (2556000,42.91) (2815600,42.78) (3101600,42.76) (3416700,42.73) (3763800,42.60) (4146100,42.55) (4567300,42.43) (5031300,42.48) (5542400,42.36) (6105400,42.32) (6725600,42.25) (7408800,42.21) (8161400,42.20) (8990400,42.12) (9903700,42.06) (10909700,42.12) (12017900,42.09) (13238800,42.04) (14583600,42.00) (16065000,42.04) (17696900,42.01) (19494600,42.04) (21474900,42.03) (23656400,41.97) (26059500,42.00) (28706600,41.98) (31622700,42.02)};
\addplot[gray!60, dashdotted] coordinates {(300,6.38) (400,7.02) (500,7.06) (600,7.44) (700,7.69) (800,7.66) (900,8.03) (1000,7.96) (1100,7.98) (1200,8.12) (1300,8.23) (1400,8.27) (1600,8.54) (1800,8.53) (1900,8.50) (2100,8.62) (2400,8.67) (2600,8.74) (2900,8.76) (3200,8.78) (3500,8.94) (3900,8.95) (4300,8.99) (4700,9.01) (5200,9.09) (5700,9.15) (6300,9.14) (6900,9.26) (7700,9.14) (8400,9.23) (9300,9.22) (10200,9.29) (11300,9.25) (12400,9.31) (13700,9.32) (15100,9.33) (16700,9.41) (18400,9.39) (20200,9.36) (22300,9.41) (24500,9.39) (27000,9.44) (29800,9.43) (32800,9.47) (36200,9.50) (39800,9.43) (43900,9.48) (48400,9.45) (53300,9.46) (58700,9.46) (64700,9.42) (71200,9.47) (78500,9.50) (86400,9.46) (95200,9.46) (104900,9.50) (115600,9.46) (127300,9.40) (140300,9.45) (154500,9.47) (170200,9.44) (187500,9.46) (206600,9.49) (227500,9.42) (250700,9.48) (276100,9.46) (304200,9.52) (335100,9.45) (369100,9.45) (406600,9.48) (447900,9.49) (493400,9.44) (543600,9.43) (598800,9.49) (659600,9.45) (726600,9.44) (800500,9.45) (881800,9.46) (971300,9.43) (1070000,9.43) (1178700,9.42) (1298500,9.41) (1430400,9.42) (1575700,9.50) (1735700,9.48) (1912100,9.50) (2106300,9.46) (2320300,9.51) (2556000,9.49) (2815600,9.46) (3101600,9.45) (3416700,9.50) (3763800,9.47) (4146100,9.50) (4567300,9.44) (5031300,9.45) (5542400,9.38) (6105400,9.43) (6725600,9.42) (7408800,9.48) (8161400,9.52) (8990400,9.45) (9903700,9.42) (10909700,9.48) (12017900,9.46) (13238800,9.49) (14583600,9.48) (16065000,9.46) (17696900,9.44) (19494600,9.46) (21474900,9.47) (23656400,9.44) (26059500,9.47) (28706600,9.43) (31622700,9.44)};
\end{axis}
\end{tikzpicture}
\end{minipage}
\par\medskip
\pgfplotslegendfromname{learnedleg}
\caption{\textbf{When the baseline must be learned.} Normalized Qini in the straddling population when every ranking is estimated from $n$ observations: baseline and threshold prioritization from $n$ untreated people; estimated BATE prioritization from an experiment with 6.6\% treated; CATE prioritization (cell estimator) from a balanced experiment. Compare \cref{fig:straddle}, where the baseline is known and sizes count treated people. Estimated threshold prioritization peaks at 92\% at intermediate sizes because it misorders two pairs of groups almost equally far from 50\%, and estimation noise sometimes reverses them into the better order.}
\label{fig:learned}
\end{figure}

% =====================================================================
\section{Details of the Criteo Illustration}\label{app:criteo}
% =====================================================================

\paragraph{Data.} We use version 2.1 of the Criteo uplift data \citep{diemert2021benchmark}: 13{,}979{,}592 users, 85\% assigned to treatment, with 12 anonymized features and binary indicators for visit, conversion, and exposure. Visit rates are 3.82\% under control and 4.85\% under treatment. Only 3.6\% of treated users were exposed to an ad; we use assignment throughout, so effects are intention-to-treat effects.

\paragraph{Splits.} A random 30\% of users (4{,}196{,}798, of whom 629{,}910 are controls) form the evaluation data, used only to evaluate rankings. The remaining 1{,}467{,}027 control users train the baseline and stay fixed. The remaining 8{,}315{,}767 treated users form the pool from which experiments are drawn without replacement.

\paragraph{Models.} All boosted models use LightGBM with a logistic loss (squared loss for the X-learner's effect models), learning rate 0.05, 63 leaves, at least 200 observations per leaf, and early stopping after 50 rounds without improvement on a random 20\% of their training data. The baseline model stops at 99 trees; its predictions $\hat\theta$ are clipped to $[10^{-6},1-10^{-6}]$, and $\hat\mu=F^{-1}(\hat\theta)$ denotes the predicted log-odds, the estimated systematic index. Baseline prioritization ranks by $\hat\theta$, from the highest down, and threshold prioritization by $-|\hat\theta-1/2|$. Given an experiment, the one-parameter logit solves $\sum_i\{Y_i-F(\hat\mu_i+\hat\delta)\}=0$ over its treated users and ranks by $-|\hat\mu+\hat\delta/2|$; the whole pool gives $\hat\delta=0.257$, and the estimate has a standard deviation across experiments of 0.16 at 1{,}000 treated users and 0.017 at 100{,}000. A more flexible BATE, a logit on a natural cubic spline in $\hat\mu$, did worse in small experiments and no better from about 300{,}000 treated users on, so we do not report it. The T-learner fits a boosted model $\hat\theta_1$ of treated outcomes on the features and ranks by $\hat\theta_1-\hat\theta$. The offset learner instead learns a shift for each user: it fits a boosted model $\hat\delta(X)$ of treated outcomes with $\hat\mu$ as its initial score and ranks by $F(\hat\mu+\hat\delta(X))-\hat\theta$. The X-learner \citep{kunzel2019metalearners} regresses $Y-\hat\theta(X)$ on the features among the treated users, giving $\hat\tau_1$, and $\hat\theta_1(X)-Y$ among the training controls, giving $\hat\tau_0$, and ranks by $\gamma\hat\tau_0+(1-\gamma)\hat\tau_1$, where $\gamma$ is the treated share of the training users.

\paragraph{Evaluation.} For each ranking, we sort evaluation users by score, breaking ties at random, and estimate $V(u)$ at $u=1\%,2\%,\ldots,100\%$ as $u$ times the difference between treated and control visit rates among the top $u$. At $u=100\%$, this is the average effect, 1.07 percentage points. We use 20 experiments of each size up to 30{,}000 treated users, 10 at 100{,}000 and 300{,}000, 5 at one million, 3 at three million, and the whole pool once, and average curves across experiments. At each capacity, crossovers use the rule of \cref{app:simdetails}: the first tested size from which the average estimated gain of CATE prioritization over estimated BATE prioritization (one-parameter logit), across repeated experiments, is positive and remains positive at every larger tested size. These are descriptive crossovers, not sizes at which gains are statistically significant. Beyond about 40\%, differences between the X-learner and estimated BATE prioritization are within $\pm0.1$ incremental visits per 1{,}000 users at every size from a million on, so crossovers there reflect noise. Because the best possible ranking is unknown, we do not normalize the Qini. \Cref{tab:criteo-qini} instead reports the mean over $u$ of $V(u)-u\,\E[\beta]$, the area between a ranking's curve and random targeting's.

\paragraph{Evaluation noise.} All rankings are evaluated on the same data, so averaging over repeated experiments does not reflect the evaluation data's sampling noise. To gauge it, we train every ranking on one experiment of a million treated users, hold these models fixed, and split the evaluation data into 20 random folds. Across folds, the standard error of the difference from estimated BATE prioritization, for any of the three CATE learners, is 0.06--0.20 incremental visits per 1{,}000 users at the top 1\%, 5\%, 10\%, 25\%, 50\%, and 80\%, and 0.06--0.12 for the Qini. For the X-learner, the differences are $+1.14$, $+1.63$, $+0.97$, $+0.13$, $-0.01$, and $-0.03$ at these fractions, and $+0.18$ (standard error 0.06) for the Qini. The other two learners' differences among the top half and beyond are two to four standard errors below zero. We do not repeat this check at other experiment sizes.

\begin{table}[ht]
\centering
\caption{\textbf{Qini in the Criteo experiment.} Mean over treatment fractions of incremental visits per 1{,}000 users beyond random targeting, by experiment size in thousands of treated users.}
\label{tab:criteo-qini}
\vspace{4pt}
\small
\begin{tabular}{lrrrrrrrrr}
\toprule
Experiment size &  1 & 3 & 10 & 30 & 100 & 300 & 1{,}000 & 3{,}000 & 8{,}316 \\
\midrule
Baseline & \multicolumn{9}{c}{3.69} \\
Threshold & \multicolumn{9}{c}{3.70} \\
BATE, one-parameter logit & 3.70 & 3.70 & 3.70 & 3.70 & 3.70 & 3.70 & 3.70 & 3.70 & 3.70 \\
CATE, T-learner & 0.48 & 1.56 & 2.06 & 2.54 & 2.74 & 3.23 & 3.58 & 3.66 & 3.69 \\
CATE, offset learner & 1.58 & 2.17 & 2.55 & 2.78 & 3.30 & 3.64 & 3.76 & 3.77 & 3.68 \\
CATE, X-learner & 2.29 & 2.97 & 2.92 & 3.16 & 3.29 & 3.76 & 3.87 & 3.95 & 3.93 \\
\bottomrule
\end{tabular}
\end{table}

\paragraph{Code.} The \texttt{code/empirical} folder contains the analysis. \texttt{run\_criteo.py} fits every ranking except the X-learner for every experiment and saves the curves (about 40 minutes on two cores); \texttt{run\_xlearner.py} does the same for the X-learner (about an hour); \texttt{criteo\_effects.py} computes the effects by baseline group and the evaluation noise; \texttt{make\_criteo\_outputs.py} prints \cref{tab:criteo,tab:criteo-qini}, the coordinates of \cref{fig:criteo}, and the numbers quoted in the text. The data must be downloaded separately.

\end{document}